\documentclass[10pt,journal,compsoc]{IEEEtran}
\usepackage{mathptmx}

\renewcommand{\baselinestretch}{1}

\usepackage{xspace}
\usepackage[usenames, dvipsnames]{color}
\usepackage{amsmath}
\usepackage{multirow}
\usepackage[pagebackref=true,breaklinks=true,letterpaper=true,colorlinks,citecolor=blue,linkcolor=blue,bookmarks=false]{hyperref}
\usepackage{paralist}
\usepackage{adjustbox}
\definecolor{gray}{rgb}{0.35,0.35,0.35}
\definecolor{MyBlue}{rgb}{0,0.2,0.8}
\definecolor{MyRed}{rgb}{0.8,0.2,0}
\definecolor{Red}{rgb}{1,0,0}
\definecolor{MyGreen}{rgb}{0.0,0.5,0.1}
\definecolor{MyGray}{rgb}{0.4,0.4,0.4}

\usepackage{graphicx}
\usepackage{subcaption}
\graphicspath{{figures/}}

\usepackage{array}
\newcolumntype{L}[1]{>{\raggedright\let\newline\\\arraybackslash\hspace{0pt}}m{#1}}
\newcolumntype{C}[1]{>{\centering\let\newline\\\arraybackslash\hspace{0pt}}m{#1}}
\newcolumntype{R}[1]{>{\raggedleft\let\newline\\\arraybackslash\hspace{0pt}}m{#1}}

\long\def\ignorethis#1{}
\def\etal{et~al.\xspace}

\newcommand{\figref}[1]{Figure~\ref{fig:#1}}
\newcommand{\tabref}[1]{Table~\ref{tab:#1}} 
\newcommand{\eqnref}[1]{(\ref{eq:#1})}
\newcommand{\secref}[1]{Section~\ref{sec:#1}}

\renewcommand{\paragraph}[1]{\textbf{#1}}

\ifCLASSOPTIONcompsoc
  \usepackage[nocompress]{cite}
\else
  \usepackage{cite}
\fi
\begin{document}
%
\title{Semantic-driven Generation of Hyperlapse\\from $360^\circ$ Video}
%
%
%
%

\author{Wei-Sheng Lai, Yujia Huang, Neel Joshi, Christopher Buehler, Ming-Hsuan Yang, and Sing Bing Kang
\\
	\IEEEcompsocitemizethanks{
		\IEEEcompsocthanksitem W.-S. Lai and M.-H. Yang are with the Department of Electrical and Engineering and Computer Science, University of California, Merced, CA, 95340. Email: $\{$wlai24$|$mhyang$\}$@ucmerced.edu
		\IEEEcompsocthanksitem Y. Huang is with the Robotics Institute, Carnegie Mellon University, PA 15213. Email: yujiah1@andrew.cmu.edu
		\IEEEcompsocthanksitem N. Joshi, C. Buehler and S. B. Kang are with Microsoft Artificial Intelligence and Research (AI\&R), Redmond, WA 98052. Email: $\{$neel$|$chbuehle$|$sbkang$\}$@microsoft.com
	}
}

%
%

\markboth{}%
{Shell \MakeLowercase{\textit{et al.}}: Bare Demo of IEEEtran.cls for Computer Society Journals}
%



\IEEEtitleabstractindextext{%
\begin{abstract}
We present a system for converting a fully panoramic ($360^\circ$) video into a normal field-of-view (NFOV) hyperlapse for an optimal viewing experience. 
Our system exploits visual saliency and semantics to non-uniformly sample in space and time for generating hyperlapses. 
In addition, users can optionally choose objects of interest for customizing the hyperlapses.
We first stabilize an input $360^\circ$ video by smoothing the rotation between adjacent frames and then compute regions of interest and saliency scores.
An initial hyperlapse is generated by optimizing the saliency and motion smoothness followed by the saliency-aware frame selection.
We further smooth the result using an efficient 2D video stabilization approach that adaptively selects the motion model to generate the final hyperlapse.
We validate the design of our system by showing results for a variety of scenes and comparing against the state-of-the-art method through a large-scale user study.
\end{abstract}

}

\maketitle

\IEEEdisplaynontitleabstractindextext

%
\IEEEpeerreviewmaketitle

\IEEEraisesectionheading{
\section{Introduction}
\label{sec:introduction}
}

\IEEEPARstart{V}{ideos} are common in social media, with apps in mobile devices making it easy to capture and share image sequences online. 
While panoramic videos are less prevalent, they are gaining popularity given the appeal of visual immersion and increasing number of commercially available cameras that capture very wide field-of-view videos (fish-eye cameras such as Kodak PixPro SP360) and $360^\circ$ videos (e.g., Ricoh Theta S, Samsung Gear 360, and Giroptic 360cam). 
Popular sites such as Youtube and Facebook have added support for viewing panoramic videos. 
There are also more sophisticated systems such as Jack-In Head~\cite{Kasahara-2015} and Jump~\cite{Anderson-TOGA-2016} that use a collection of cameras for capturing omni-directional stereo videos, but these are designed for virtual reality (VR) devices and are cost prohibitive.

In this paper, we address the difficult issues of viewing $360^\circ$ videos. 
Unlike a normal field-of-view (NFOV) video, a single frame from a $360^\circ$ video contains considerable distortion after the equirectangular projection. 
In addition, not all parts of the scene in space and time are equally interesting. 
There are available tools to render part of the frame while the video is played, but the view is manually selected either through mouse control or by orienting the mobile device. 
The timeline is typically modified through manual interaction with the time progress slider. 
This may not be the optimal viewing experience, given that important or interesting content may be missed, and it can take too much time to watch a long video.

One solution to reduce the burden of watching long videos is to automatically speed up the sequences while preventing visual jerkiness through stabilization. 
The resulting video, called the hyperlapse, can be generated using a number of techniques~\cite{Kopf-TOG-2014,Joshi-TOG-2015}. 
However, these methods operate on NFOV videos and the changes to viewpoint are minimal. 
In addition, existing methods that compute the virtual camera paths typically do not consider semantic information and user preferences when generating hyperlapses.

\begin{figure*}
	\centering
	\includegraphics[width=0.90\textwidth]{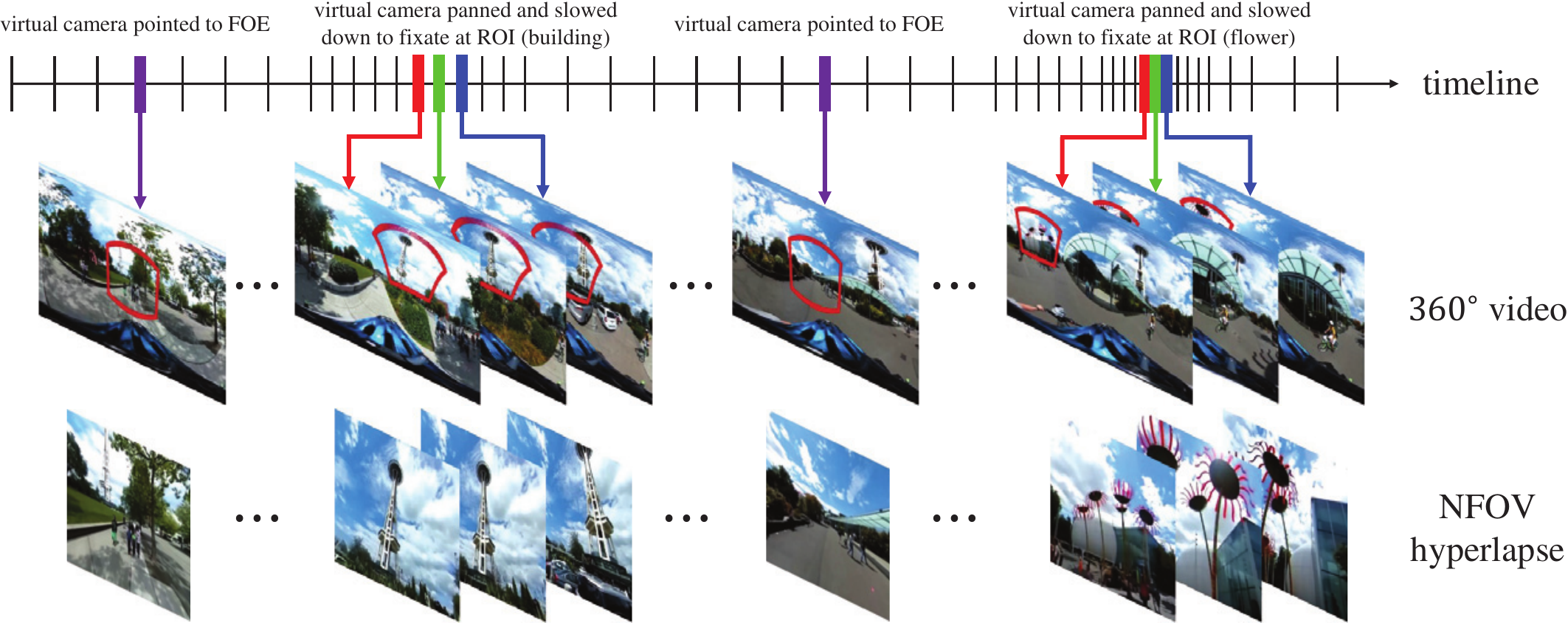}
	\caption{Our system converts a fully panoramic (i.e., $360^\circ$) video into a normal field-of-view hyperlapse that is optimized for an viewing experience.
		We exploit visual saliency and semantics in a $360^\circ$ video to identify objects of interest for planning a camera path and rendering a hyperlapse. 
		In addition, users can select objects of interest to generate hyperlapses tailored for individual preferences.}
	\label{fig:teaser}
\end{figure*}

We propose to optimize the viewing experience of watching a long $360^\circ$ video by using visual saliency and scene semantics to generate an NFOV hyperlapse. 
The saliency score of an area in a scene is based on the relative motion and appearance of objects.
In addition, object categories are determined based on a state-of-the-art segmentation method that provides semantic information. 
A user first selects the speed-up rate of the output video with the option of customizing the hyperlapse by indicating objects of great interest.
Our system then computes an optimal camera path in space and time by considering saliency scores, semantic preferences, and temporal smoothness over the video. 
Finally, we adopt an efficient 2D video stabilization approach that adaptively selects per-frame motion models to render a smooth hyperlapse.
\figref{teaser} depicts the overview of the proposed approach that generated NFOV hyperlapses from $360^\circ$ videos.

We show examples of hyperlapses from $360^\circ$ videos in various scenes with comparisons to existing methods. 
%
We carry out a user study to evaluate the proposed system with different design options and compare with the 
state-of-the-art technique (Pano2Vid~\cite{pano2vid} with speed-up using the hyperlapse approach of Joshi~\etal~\cite{Joshi-TOG-2015}). 
Results of the user study indicate that more subjects prefer hyperlapses generated from our approach.

The contributions of this work are summarized as follows:
\begin{itemize}
	\item We develop the first system to generate an NFOV hyperlapse from a $360^\circ$ video. 
	(We note that the output of the Pano2Vid method~\cite{pano2vid} is of the same length as the input video and hence not a hyperlapse.)
	\item We propose a semantic-driven approach that facilitates non-uniform sampling in space and time for optimal panning and fixating on objects of interest. 
	%
	\item We present a robust 2D stabilization method that adaptively selects the motion model for generating smooth hyperlapses. 
\end{itemize}

\section{Related Work}
In this section, we review the most relevant work on $360^\circ$ video and hyperlapse. 
We also review recent methods on video stabilization and spatial-temporal saliency detection, as these are integral components of our system.

\subsection{Converting $\mathbf{360^\circ}$ Videos to NFOV Videos}
The work closest to ours is the Pano2Vid method by Su~\etal~\cite{pano2vid} that generates NFOV sequences from $360^\circ$ videos. 
This method learns a capture-worthiness score for each predetermined spatial-temporal segment and uses a dynamic programming algorithm to select a smooth camera path for rendering an NFOV video.
Our method differs from the Pano2Vid method in four aspects:
\begin{itemize}
	\item Given a $360^\circ$ video, our system renders a hyperlapse whereas the Pano2Vid method generates an NFOV video of the same length as the input.
	%
	\item 
	The Pano2Vid method uses the holistic video contents for learning the capture-worthiness scores, 
	%
	while we combine low-level spatial-temporal visual saliency with high-level image semantics to detect interesting areas or objects.
	Our system provides users with more options (i.e., choosing semantic labels of individual preferences) to customize the hyperlapses.
	%
	%
	\item Our algorithm optimizes the scene semantics and visual camera movements (direction of forward motion, panning speed and acceleration) for determining virtual camera paths.
	The direction of forward motion (i.e., focus of expansion) provides more understanding of the scenes and activities in the videos, which are not considered in the Pano2Vid method.
	%
	\item We stabilize the output videos using sparse feature trajectories, while the Pano2Vid method controls the smoothness of cameras with a spatial penalty term when solving a dynamic programming problem. 
	Since the Pano2Vid method does not directly stabilize image content, the output video may retain the visual jitter that is originally in the input.
\end{itemize}

\subsection{Rendering Hyperlapses}
\label{sec:related_hyperlapse}
Numerous approaches have been developed to convert an NFOV video to a hyperlapse.
The Instagram hyperlapse app first uniformly skips frames and then stabilizes videos using sensor data from the inertial measurement unit (IMU)~\cite{Karpenko-CSTR-2011}. 
%
As such, it cannot be used to generate a hyperlapse from a pre-recorded video.
Kopf~\etal~\cite{Kopf-TOG-2014} propose a 3D approach based on the structure-from-motion (SfM) and image-based rendering to compute a smooth 6D camera path for generating a hyperlapse.
Although it can generate smooth videos for scenarios with significant parallax,
the SfM operation is computationally expensive and may fail when the camera translation is small, or there are significant amounts of pure rotation.

The 2D approaches are generally more efficient and robust than 3D methods but do not account for parallax.
Poleg~\etal~\cite{Poleg-CVPR-2015} and Ramos~\etal~\cite{Ramos-ICIP-2016} construct a graph for the entire video and find the shortest path to sample frames for generating a hyperlapse.
Joshi~\etal~\cite{Joshi-TOG-2015} propose an adaptive selection strategy to sample a set of optimal frames that minimize the alignment errors and then smooth the results using an efficient 2D video stabilization method.
This approach is capable of performing in real-time on mobile devices.
However, the above-mentioned approaches are based on 2D motion analysis such as optical flow and homography transformations, which cannot be directly applied to $360^\circ$ videos.
Kopf~\cite{Kopf-TOGA-2016} creates a $360^\circ$ hyperlapse by first stabilizing the input video and then dropping frames based on the estimated camera velocity.

The scope of our work differs from these methods in that we create \textit{NFOV} hyperlapses from \textit{360}$^\circ$ videos.
We summarize the main differences between the above-mentioned hyperlapse approaches, Pano2Vid, and our system in~\tabref{hyperlapse_compare}.

\begin{table*}[t]
	\centering
	\caption{
		Comparisons of existing hyperlapse approaches, Pano2Vid~\protect\cite{pano2vid} and the proposed algorithm.
		We note that the Pano2Vid method does not speed up the videos to generate hyperlapses.
		%
		The Pano2Vid method uses a data-driven approach to learn the capture-worthiness scores based on holistic contents, while our algorithm combines low-level spatial-temporal saliency with high-level semantic segmentation for detecting interesting regions or objects in videos, which facilitates camera planning for better viewing experience. 
	}
	\vspace{-2mm}
	\begin{tabular}{rcccccc}
		\hline
		\textcolor{blue}{Method} & 
		\textcolor{blue}{Input} &
		\textcolor{blue}{Output} &
		\textcolor{blue}{Speed-up} &
		\textcolor{blue}{Camera panning} &
		\textcolor{blue}{Saliency/Semantic} &
		\textcolor{blue}{Stabilization} \\
		\hline\hline
		\textcolor{Brown}{Kopf~\etal~\cite{Kopf-TOG-2014}} & 
		NFOV video & NFOV video & 
		Yes & No & No & 3D \\
		\textcolor{Brown}{Joshi~\etal~\cite{Joshi-TOG-2015}} &
		NFOV video & NFOV video & 
		Yes & No & No & 2D \\
		\textcolor{Brown}{Poleg~\etal~\cite{Poleg-CVPR-2015}} &
		NFOV video & NFOV video & 
		Yes & No & No & No \\
		\textcolor{Brown}{Ramos~\etal~\cite{Ramos-ICIP-2016}} &
		NFOV video & NFOV video & 
		Yes & No & Face / Pedestrian & No \\
		\textcolor{Brown}{Kopf~\cite{Kopf-TOGA-2016}} &
		$360^\circ$ video & $360^\circ$ video & 
		Yes & No & No & Hybrid 3D-2D\\
		\textcolor{Brown}{Pano2Vid~\cite{pano2vid}} &
		$360^\circ$ video & NFOV video & 
		No & Yes & Yes & No \\
		\textcolor{magenta}{Ours} &
		$360^\circ$ video & NFOV video & 
		Yes & Yes & Yes & $360^\circ$ + 2D\\
		\hline
	\end{tabular}
	\label{tab:hyperlapse_compare}
	\vspace{-0.2cm}
\end{table*}

\subsection{Video Stabilization}
There is a considerable body of work on video stabilization, and we discuss the most relevant methods in this section.
Stabilization algorithms can be categorized into 3D and 2D approaches.
3D methods~\cite{Buehler-CVPR-2001,Liu-TOG-2009} often reconstruct the scene geometry and smooth the 6D camera poses to stabilize videos.
Although these approaches are able to handle parallax, the effectiveness of the 3D reconstruction process depends highly on camera motion as discussed in~\secref{related_hyperlapse}.
In general, 3D approaches tend to be less robust than 2D methods.
In~\cite{Liu-TOG-2011}, Liu~\etal introduce smoothness constraints on the subspace of feature trajectories, which is a hybrid 2D and 3D approach.

2D stabilization algorithms (e.g., \cite{Matsushita-PAMI-2006}) are developed based on simple motion models such as similarity or homography transformations.
However, these methods are not effective for handling scenes with significant parallax.
Grundmann~\etal~\cite{Grundmann-CVPR-2011} constrain the camera path as a concatenation of constant, linear, and parabolic motions by solving an $L_1$ optimization problem.
Recent approaches use more flexible motion models, such as mesh-warping and bundled camera paths~\cite{Liu-TOG-2013}, to handle parallax and rolling shutter effects.
%
The computational cost of optimizing these flexible motion models increases as a result.
In addition, these models have to be carefully regularized such that the stabilized videos do not contain temporal artifacts (e.g., shearing and wobbling).
%
Note that approaches based on 2D motion models cannot be directly applied to $360^\circ$ videos.

Several methods have been recently developed for handling videos with wide viewing angles. 
Kasahara~\etal~\cite{Kasahara-2015} estimate and smooth relative rotation between adjacent frames to stabilize $360^\circ$ videos.
On the other hand, Kamali~\etal~\cite{Kamali-2011} propose a 3D approach based on structure-from-motion and spherical image warping.
Most recently, the approach proposed by Kopf~\cite{Kopf-TOGA-2016} uses a deformable rotation motion model for $360^\circ$ video stabilization.

In this work, videos are stabilized in two stages. 
We first stabilize the input $360^\circ$ video to achieve better temporal coherence for video content analysis.
After rendering an NFOV video from a $360^\circ$ input, we apply a 2D stabilization method to improve temporal smoothness.
We observe that the conventional 2D stabilization approach based on homographies may fail during camera panning or rapid scene changing in a hyperlapse.
%
%
For robustness, we adaptively select motion models from translation, similarity, and homography transformations based on the Akaike information criterion (AIC)~\cite{AIC,Gheissari-2003}.
%
We show that this adaptive stabilization approach generates stable hyperlapses with fewer temporal artifacts in~\secref{results}.

\subsection{Visual Saliency and Semantics}
Content-aware video re-targeting techniques~\cite{Wang-TOG-2010,Wang-TOG-2011} require importance maps to preserve visual information while solving optimization problems.
The importance maps are generally based on a combination of low-level (e.g., image gradient and optical flow) and high-level cues (e.g., visual saliency~\cite{Itti-PAMI-1998} and detected faces).
Zhou~\etal~\cite{Zhou-CVPR-2014} propose a space-time saliency method based on appearance and motion contrast as well as statistical priors that reflect the location and foreground probability. 
However, this method assumes that the camera is static and salient objects appear closer to the image center.
In contrast, we make no such assumptions in this work as salient objects are likely to appear anywhere in a $360^\circ$ view with equal probabilities.

Semantic segmentation methods~\cite{FCN,Zheng-ICCV-2015} based on deep convolutional neural networks (CNNs) have been recently developed for labeling objects in the scenes.
Since we do not require highly accurate semantic segmentation, we use the frame-based fully convolutional network (FCN)~\cite{FCN} instead of video-based schemes~\cite{Kundu-CVPR-2016} for the efficiency reason.
We then aggregate the probabilities and assign a semantic label for each spatial-temporal region in a video.

The approach of Ramos~\etal~\cite{Ramos-ICIP-2016} embeds semantic information when generating hyperlapses from NFOV videos.
This method uses semantic scores based only on the face or pedestrian detection to adaptively adjust the speed of a hyperlapse in frame selection.
%
The visual semantics considered in this work are significantly richer as the FCN is able to label 60 classes (from the Pascal-Context dataset~\cite{pascal-context}).
As a result, more interesting objects can be identified. 
In addition to operating on $360^\circ$ videos, our system provides more options for users to customize a hyperlapse.


\begin{figure*}[t]
	\centering
	\includegraphics[width=\textwidth]{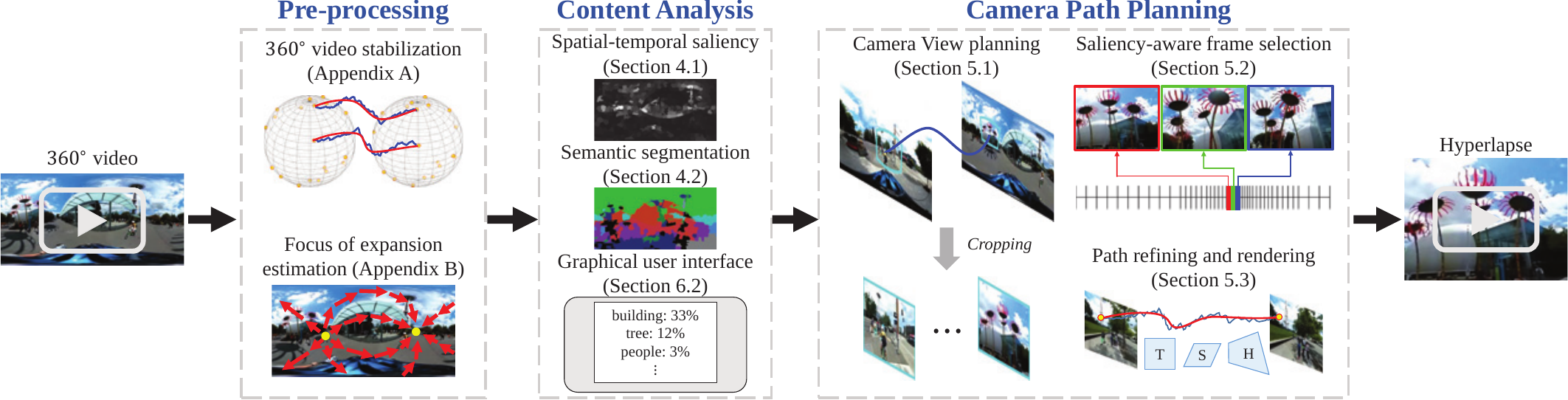}
	\caption{Pipeline of the proposed algorithm. 
		Given a $360^\circ$ video, we first stabilize the sequence to smooth the relative rotation between adjacent frames.
		We estimate the focus of expansion (i.e., the direction of forward motion) as a prior information for our camera path planning.
		To extract the regions of interest, we compute the spatial-temporal saliency and semantic segmentation.
		The detected regions of interest are used to guide the camera path planning.
		Finally, we use an adaptive 2D video stabilization to render a smooth hyperlapse.
	}
	\vspace{-2mm}
	\label{fig:overview}
\end{figure*}

\section{System Overview}
The pipeline of our system is shown in~\figref{overview}.
%
Most $360^\circ$ videos are captured with hand-held cameras and thus typically shaky.
It is necessary to stabilize an input video before analyzing the contents.
%
We first track a set of feature trajectories using the KLT method~\cite{KLT} and estimate the relative rotation between adjacent frames.
%
We then smooth and interpolate the rotations with a Gaussian function and warp the input frames using the corrective 3D rotations (Appendix A).

After stabilizing the input video, we find the forward camera motion by estimating the focus of expansion (FOE), which is used as a prior in camera path planning. 
The FOE is computed using the optical flow and Hough transform (Appendix B).
%

To exploit image semantics for camera path planning, we extract regions of interest (ROIs) using spatial-temporal saliency detection (\secref{saliency_detection}) and semantic segmentation (\secref{semantic_segmentation}).
Our graphical user interface shows parsed semantic information, which provides users with the options for selecting objects of interest and modifying virtual camera settings such as speed and field-of-view (\secref{GUI}).

Our path planning algorithm utilizes information from the regions of interest, the focus of expansion, and user settings.
It computes a virtual camera path by first optimizing the camera viewing direction across the entire video (\secref{view_planning}).
The NFOV video is extracted by performing perspective projection and cropping from the $360^\circ$ video. 
This is followed by assigning an importance score to each frame and selecting the optimal frames (\secref{frame_selection}).
The output is produced by refining the initial NFOV hyperlapse using a 2D stabilization method with adaptive motion model selection (\secref{final_stabilization}).


\section{Video Content Analysis}
\label{sec:ROI_detection}
Since our goal is to generate a semantic-driven hyperlapse that highlights interesting objects, we analyze the video and associate regions with semantic labels. 
We use a semantic segmentation method to extract object labels in each frame.
The object labels are then combined with visual and motion saliency scores to generate regions of interest, 
which are used to guide the camera path planning.
%
If no dominant ROIs are detected, the viewing direction is encouraged to be close to the forward camera motion as measured by FOE.

\subsection{Visual Saliency}
\label{sec:saliency_detection}
We first over-segment an input video into temporal superpixels (TSPs)~\cite{TSP} and adopt a bottom-up method similar to~\cite{Zhou-CVPR-2014} to compute the spatial-temporal saliency scores for each TSP.
A spatial-temporal region is considered salient if its color or motion is different from its neighbors.
We thus measure the feature contrast based on appearance and motion statistics.

We denote the TSP centered at location $c$ in frame $t$ as $r_{c,t}$.
For each TSP $r_{c,t}$, we extract three feature vectors: color histogram in the CIE-Lab color space, $\mathbf{x}^{\text{col}}_{c,r}$, 
histogram of flow magnitude, $\mathbf{x}^{\text{mag}}_{c,r}$, and 
histogram of flow orientation, $\mathbf{x}^{\text{ori}}_{c,r}$.
Each feature vector is normalized to be of unit length.
We define the feature of $r_{c,t}$ as the concatenation of these three feature vectors:
\begin{equation}
\mathbf{x}_{c,r} = [ \mathbf{x}^{\text{col}}_{c,r} \; ; \; \mathbf{x}^{\text{mag}}_{c,r} \; ; \; \mathbf{x}^{\text{ori}}_{c,r} ].
\end{equation}
The feature contrast of each TSP is measured as the weighted sum of its feature distance to 
the other TSPs:
\begin{equation}
s_{c,t} = \sum_{r_{i,t} \neq r_{c,t}} |r_{c,t}| \cdot w(r_{i,t}, r_{c,t}) \cdot \| \mathbf{x}_{i,t} - \mathbf{x}_{c,t} \|^2,
\end{equation}
where $w(r_{i,t}, r_{c,t}) = \exp\left( -\| \mathbf{m}_{i,t} - \mathbf{m}_{c,t} \| /  \sigma_s \right)$ is the weight between the center of mass, $\mathbf{m}_{i,t}$ and $\mathbf{m}_{c,t}$ of the TSPs $r_{i,t}$ and $r_{c,t}$, respectively.
We compute the feature contrast $s_{c, t}$ within a sliding temporal window of 200 frames.
We normalize the pixel coordinate to $[0, 1]$ and set $\sigma_s$ to $0.04$.
We use the feature contrast as the saliency score of a TSP.

\subsection{Semantic Segmentation}
\label{sec:semantic_segmentation}
\begin{figure}[t]
	\centering
	\begin{tabular}{cc}
		\includegraphics[width=0.45\columnwidth]{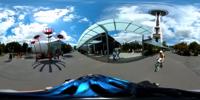} &
		\includegraphics[width=0.45\columnwidth]{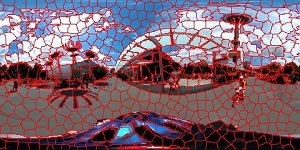} \\
		(a) Input frame 		&
		(b) TSP segmentation 	
		\\
		\includegraphics[width=0.45\columnwidth]{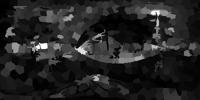} &
		\includegraphics[width=0.45\columnwidth]{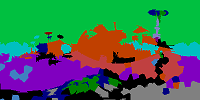} \\
		(c) Saliency map 		&
		(d) Semantic labels
	\end{tabular}
	\caption{Video content analysis. 
		First, we segment an input video into several temporal superpixels and compute the spatial-temporal saliency scores using appearance and motion features.
		We perform the semantic segmentation and then assign a semantic label and a saliency score to each TSP.
	}
	\vspace{-2mm}
	\label{fig:ROI_detection}
\end{figure}

We apply FCN~\cite{FCN} to each frame independently to obtain the initial semantic labels.
We use the FCN-8s model pre-trained on the Pascal-Context dataset~\cite{pascal-context}, which contains 60 categories of common objects in the wild.
The initial per-frame label maps are noisy and not temporally coherent.
We then compute the mean probability of each class within a TSP and assign the semantic label with the maximum probability for each TSP.
As a result, every TSP in the input video has a saliency score and a corresponding semantic label.
We show examples of the TSP segmentation, saliency map and semantic labels of a $360^\circ$ video frame in~\figref{ROI_detection}.

The regions of interest in an input video are selected from the top-$k$ scores of the chosen labels.
By default, our algorithm selects the label with the highest cumulative saliency score.
A user has the option to customize the output hyperlapse by manually selecting the preferred labels from our graphical user interface (see~\secref{GUI}).
%
In our experiments, we partition an input video into sub-sequences of 2,000 frames to alleviate the memory issue when computing TSPs.
For each sub-sequence (about 33 seconds), we choose 3 TSPs with the maximum scores as the detected ROIs.
%

\begin{figure*}[t]
	\centering
	\begin{tabular}{cc}
		\includegraphics[width=0.48\textwidth]{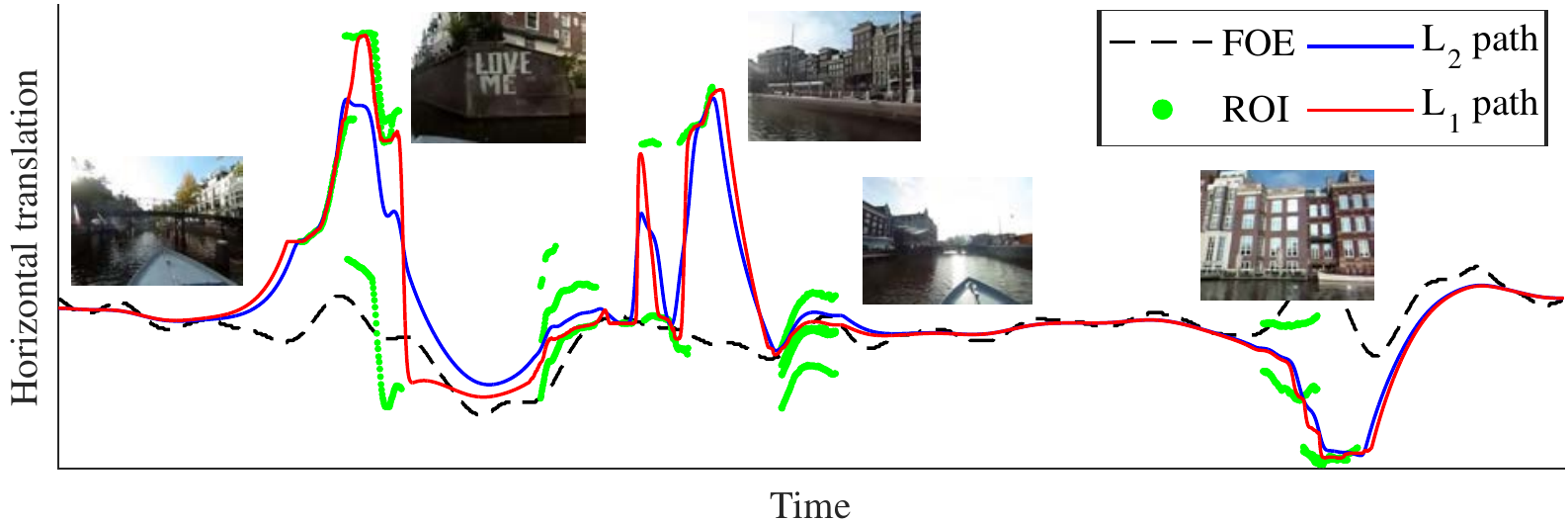} 
		&
		\includegraphics[width=0.48\textwidth]{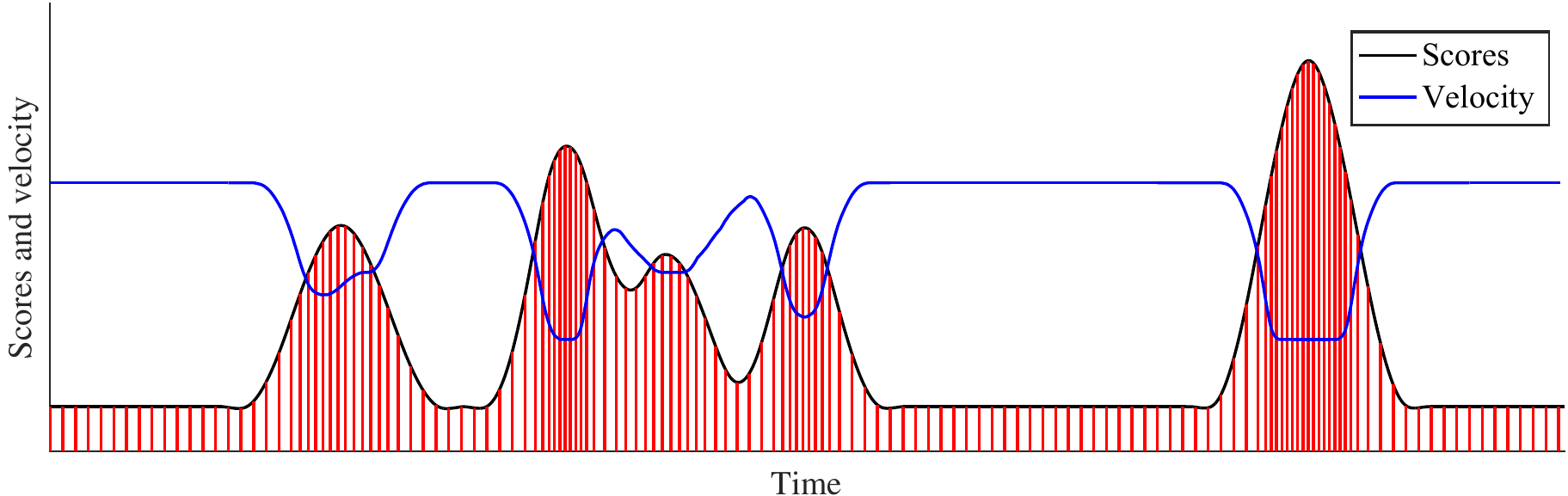} \\
		(a) View planning & (b) Saliency-aware frame selection
	\end{tabular}
	\caption{Examples of a camera path computed from our path planning algorithm.
		(a) Our view planning algorithm optimizes the viewing directions across the entire sequence.
		(b) We adopt a saliency-aware frame selection approach to select a set of optimal frames.
	}
	\vspace{-3mm}
	\label{fig:path_planning}
\end{figure*}


\section{Camera Path Planning for Hyperlapse}
\label{sec:path_planning}
%
In this section, we describe the proposed algorithm to compute a virtual camera path from an input $360^\circ$ video, which consists of determining NFOV viewpoints at every frame and selecting a subset of frames. 
A virtual camera path in a $360^\circ$ video is a set of camera poses $\mathbf{p}_t = (\theta_t, \phi_t)$ that indicates the viewing direction $(\theta_t, \phi_t)$ at frame $t$.
We assume that the head vector of the virtual camera always points to the top of the 3D sphere, i.e., we do not allow any rotation around the axis from the sphere center pointing at the center of  NFOV videos.
%
To generate a hyperlapse, the sequence of time samples should be monotonically increasing and close to the target speed-up rate, e.g., $8$ or $16$ times.
To make the path planning problem (i.e., estimating $\theta$, $\phi$ and $t$) more tractable, we propose a three-phase method:
\begin{enumerate}
	\item \textbf{View planning}: Given regions of interest as guidance and focus of expansion as a prior, we optimize the camera viewing direction for each frame.
	\item \textbf{Frame selection}: We select the optimal set of frames as a trade-off among saliency scores, target speed-up rate, temporal acceleration, and frame-to-frame alignment errors.
	\item \textbf{Path refinement and rendering}: Given one selected path, we determine the amount of zooming and stabilize and render a smoothed NFOV hyperlapse.
\end{enumerate}

\subsection{View Planning}
\label{sec:view_planning}
Given the detected ROIs, we first decompose each ROI into multiple camera poses.
We denote the region covered by an ROI at frame $i$ as a set of coordinates ${(\theta_i^k, \phi_i^k)}, \forall k \in [1, N]$ where $N$ is the number of pixels within the ROI.
We define the camera pose of an ROI at frame $i$ by its center of mass $p_i = (\theta_i^m, \phi_i^m)$, where:
\begin{equation}
	\theta_i^m = \frac{\sum_{k=1}^{N}\theta_i^k}{N}, \phi_i^m = \frac{\sum_{k=1}^{N}\phi_i^k}{N}.
\end{equation}
If an ROI appears from frame $t_s$ to $t_e$, we decompose the ROI into $(t_e - t_s + 1)$ camera poses.
%
%
We denote the collection of camera poses from all ROIs as $\mathbf{p}^{ROI}$.
Given the regions of interest $\mathbf{p}^{ROI}$ and focus of expansion $\mathbf{p}^{FOE}$, the smoothed camera path is extracted by minimizing the following cost function:
\begin{align}
\sum_{t=1}^{T} 
w_{r} C_{r}(\mathbf{p}_t; \mathbf{p}^{ROI}) 
+ w_{f} C_{f}(\mathbf{p}_t; \mathbf{p}^{FOE}) 
+ w_{v} C_{v}(\mathbf{p}_t) 
+ w_{a} C_{a}(\mathbf{p}_t),
\label{eq:view_planning}
\end{align}
where $T$ is the total number of input frames and 
\begin{align}
C_{r}(\mathbf{p}_t; \mathbf{p}^{ROI}) &= \sum_{i \in \Omega_t} \tilde{w}_{ti} s_i \| \mathbf{p}_t - \mathbf{p}_{i}^{ROI} \|_1, \\
C_{f}(\mathbf{p}_t; \mathbf{p}^{FOE}) &= \| \mathbf{p}_t - \mathbf{p}_{t}^{FOE} \|_2^2, \\
C_{v}(\mathbf{p}_t) &= \| \mathbf{p}_t - \mathbf{p}_{t-1} \|_2^2, \\
C_{a}(\mathbf{p}_t) &= \| \mathbf{p}_{t+1} - 2 \mathbf{p}_t + \mathbf{p}_{t-1} \|_2^2.
\end{align}
The data term $C_r$ encourages the camera viewing direction towards interesting objects or regions in a video.
The weight $\tilde{w}_{ti} = \exp(-(t - i)^2 / \sigma_t^2 )$ is defined by the time difference between the current path and ROIs.
In addition, $\sigma_t$ controls the temporal smoothness of camera panning.
$s_i$ is the saliency score of the ROI and $\Omega_t$ is the neighbors of frame $t$.
We set $\Omega_t = [t - 3 \sigma_t, t + 3 \sigma_t]$ in our experiments.
Note that we minimize the $L_1$-norm instead of $L_2$-norm in the data term.
The solution of $L_2$-norm is easily affected by outliers, e.g., ROIs with small saliency scores.
If there exist multiple ROIs in the same frame, minimizing the $L_2$-norm may lead to an average path that is not close to any ROI.
On the contrary, the $L_1$-norm is more robust and produces stable path that balances the data term and smoothness terms in~\eqnref{view_planning}.
%
%
In~\figref{path_planning}(a), we show the paths obtained by optimizing the $L_2$-norm (blue curve) and $L_1$-norm (red curve) in the data term.

The second term $C_f$ is the prior that enforces the camera path to be close to the FOE if there are no interesting ROIs in the current frame.
In addition, $C_v$ and $C_a$ are the velocity and acceleration terms that control the path smoothness.
We use the iterative re-weighted least squares (IRLS) method to optimize~\eqnref{view_planning}.
We compute an initial solution by replacing the $L_1$-norm with $L_2$-norm.
%
%
At each iteration, we formulate the problem as a least squares optimization and solve it using the conjugate gradient method.
In this work, we empirically choose the weights $w_r = 5$, $w_f = 1$, $w_v = 50$ and $w_a = 10$ for all experiments.
Furthermore, we set $\sigma_t = 10 \bar{v}$ where $\bar{v}$ is the target speed-up rate of the hyperlapse (e.g., we use $\bar{v} = 4$ or 8 in most testing videos, see~\secref{dataset}).
%

\subsection{Saliency-aware Frame Selection}
\label{sec:frame_selection}
%
%
Given an input video with a total of $T$ frames, the next step is to find a mapping from the output to input time stamp $f(\tilde{t}) = t$ where $t \in [1, T]$, $\tilde{t} \in [1, \tilde{T}]$ and $\tilde{T}$ is the number of frames in the output video.
The time difference between subsequent output frames should be close to the target speed-up rate $\bar{v}$, i.e., $f(\tilde{t}) - f(\tilde{t}-1) \approx \bar{v}$.
%
%
In addition, the frames should be shown at a slower rate when the camera is panned toward more interesting regions.
%
To avoid motion sickness caused by rapid panning and to ensure interesting regions can be easily viewed, it is crucial to slow down the speed when the camera viewing directions are changed.

Given the viewing direction $\mathbf{p}_t = (\theta_t, \phi_t), \forall t \in [1,T]$, we first apply the perspective projection to render an NFOV video with a fixed field-of-view (e.g., $100^\circ$).
We then use a variant of the frame selection algorithm described in Joshi~\etal~\cite{Joshi-TOG-2015} to select a set of optimal frames by considering saliency scores, frame alignment errors, speed, and acceleration penalties.

{\flushleft \bf{Saliency cost.}}
In order to determine the hyperlapse speed change when the camera is panned, we assign an importance score for each frame.
The importance score $s_t$ of the frame $t$ is assigned from the saliency score of the ROI closest to the current camera center.
We only assign the importance score when ROIs are detected in the current frame.
To equally distribute the importance scores in the output video, the cumulative score between successive frames should be close to a constant $\bar{v} \cdot \bar{s}$, where $\bar{s}$ is the average importance score and $\bar{v}$ is the target speed-up rate.
We thus define a function that computes saliency cost between frame $i$ and $j$:
%
\begin{equation}
L_s(i, j) = \left(\sum_{p = i}^j s_p - \bar{v} \cdot \bar{s} \right)^2.
\label{eq:saliency_cost}
\end{equation}
%
For the frames with higher scores, the proposed algorithm skips less frames and thus slows down the speed.
We show an example of the importance scores, frame sampling, and corresponding hyperlapse speed (i.e., temporal difference between selected frames) in~\figref{path_planning}(b).

{\flushleft \bf{Frame alignment cost.}}
The frame alignment cost is defined to penalize the alignment error between frames~\cite{Joshi-TOG-2015}.
We re-use the KLT features detected in the $360^\circ$ video stabilization stage.
We denote $\mathbf{x}_p^t = [x_p, y_p, 1]_t^T$ as the $p$-th feature in frame $t$ and $\mathbf{H}(i, j)$ as the homography that warps feature points from frame $i$ to $j$.
The frame alignment cost function is defined as:
\begin{equation}
L_m(i, j) = 
\begin{cases}
L_o(i, j) 	& \text{, if } L_r(i, j) < \tau_m\\
\gamma    	& \text{, if } L_r(i, j) \geq \tau_m
\end{cases},
\label{eq:alignment_cost}
\end{equation}
where $L_r$ is the re-projection error of the $n$ matched feature points between frame $i$ and $j$:
\begin{equation}
L_r(i, j) = \frac{1}{n} \sum_{p=1}^{n} \| \mathbf{x}_p^j - \mathbf{H}(i, j)\mathbf{x}_p^i \|^2.
\label{eq:reprojection_cost}
\end{equation}
In addition, $L_o$ measures the translation of the image center $\mathbf{x}_o$:
\begin{equation}
L_o(i, j) = \| \mathbf{x}_o^j - \mathbf{H}(i, j)\mathbf{x}_o^i \|^2.
\label{eq:center_cost}
\end{equation}
We set $\tau_m = 0.1 d$ and $\gamma = 0.5 d$ where $d$ is the image diagonal in pixels.
If the feature re-projection error~\eqnref{reprojection_cost} is low, the cost is equal to the motion cost~\eqnref{center_cost}.
However, if the re-projection error is too large, the estimated transformation is not reliable and we truncate the cost to a large constant.

{\flushleft \bf{Velocity and acceleration costs.}}
To achieve the target speed-up rate $\bar{v}$, we adopt a velocity cost term from~\cite{Joshi-TOG-2015}:
\begin{equation}
L_v(i, j) = \min\left( \|(j-i) - \bar{v}\|^2, \tau_v \right),
\end{equation}
which is a truncated $L_2$-norm based on the difference between the jump from frame $i$ to $j$ and the target speed-up rate $\bar{v}$.
%
To avoid sudden speed change, we compute the acceleration penalty to jump from frame $h$ to $j$ via $i$:
\begin{equation}
L_a(h,i,j) = \min\left( \|(j - i) - (i - h) \|^2, \tau_a \right).
\end{equation}
We empirically choose $\tau_v = 200$ and $\tau_a = 200$.

{\flushleft \bf{Optimal frame selection.}}
The overall cost function of the frame selection is defined as:
%
\begin{equation}
L(h, i, j) = L_m(i, j) + w_s L_s(i, j) + w_v L_v(i, j) + w_a L_a(h, i, j).
\label{eq:frame_selection_cost_all}
\end{equation}
Empirically the algorithm performs well with the following parameters: $w_s = 5000$, $w_v = 200$ and $w_a = 100$. 
We use a relatively large value for $w_s$ to emphasize the effect of the saliency cost and balance the numerical range among other terms.
We determine the optimal set of frames by solving a mapping $f^{\ast}$ in the following cost function:
\begin{equation}
f^{\ast} = \arg\min \sum_{\tilde{t}=1}^{\tilde{T}-1} L\left(f(\tilde{t}-1), f(\tilde{t}), f(\tilde{t}+1)\right).
\label{eq:frame_selection_cost_solve}
\end{equation}
%
%
Minimizing~\eqnref{frame_selection_cost_solve} is a discrete optimization problem.
We use the dynamic time warping (DTW) algorithm~\cite{Joshi-TOG-2015} to solve~\eqnref{frame_selection_cost_solve} and find the optimal set of frames.

%
%

\subsection{Path Refinement and Rendering}
\label{sec:final_stabilization}
Once the optimal frames are selected, we then determine the camera zoom and stabilize the rendered video.
We allow the camera to gradually zoom in to focus on interesting objects and resume to the default FOV afterward.
With the computed FOV curve and the selected frames, we re-render an NFOV video from the $360^\circ$ video.
Although we minimize the frame alignment errors when selecting the optimal frames, the NFOV video may not be sufficiently smooth due to frame skipping.
Therefore, we further smooth the rendered video using an efficient 2D video stabilization method with adaptive model selections.

\begin{figure}[t]
	\centering
	\begin{tabular}{cc}
		\includegraphics[width=0.48\columnwidth]{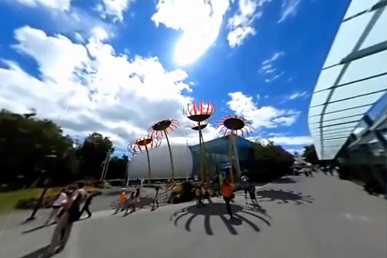} 
		\hspace{-4mm} &
		\includegraphics[width=0.48\columnwidth]{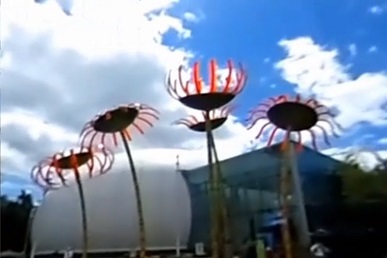} \\
		(a) Without zoom-in 
		\hspace{-4mm} & 
		(b) With zoom-in
	\end{tabular}
	\vspace{-2mm}
	\caption{Rendered frames with and without zooming effect. 
		We determine the amount of zoom-in according to the size of interesting regions.
		Users can adjust the zoom-in from our GUI as well.}
	\label{fig:zoom}
	\vspace{-4mm}
\end{figure}

{\flushleft \bf{Zooming effect.}}
\label{sec:zooming}
We determine the amount of zoom-in according to the size of interesting objects.
The horizontal FOV (in degrees) of frame $t$ can be written as $f_t = 360 \cdot W_o / W_i$, where $W_i$ is the width of the input $360^\circ$ frame and $W_o$ is the maximal width of the cropped view.
%
%
We denote the size of an ROI in the frame $t$ as $A_t$, 
and define the ratio between $A_t$ and the size of the view cropped from the $360^\circ$ frame as:
\begin{equation}
r_t = \frac{A_t}{W_o H_o} = \frac{c A_t}{W_o^2},
\label{eq:ROI_ratio}
\end{equation}
where $W_o$ and $H_o$ are the maximal width and height of the cropped view and $c = W_o / H_o$ is the aspect ratio.
The above equation is an approximation as the cropped view is not a square image (see~\figref{teaser}).
By substituting $W_o$ with $\sqrt{cA_t / r_t}$ from~\eqnref{ROI_ratio}, $f_t$ can be written as:
\begin{equation}
f_t = \sqrt{\frac{c A_t}{r_t}} \cdot \frac{360}{W_i}.
\label{eq:ft_r_t}
\end{equation}
As such, a larger $r_t$ results in a smaller $f_t$ and leads to larger camera zoom-in.
To make the FOV change gradually, we apply a Gaussian filter to smooth the FOV curve.
According to the FOV curve and camera path, we re-render an NFOV video with the zooming effect.
We fix the aspect ratio $c$ to $4/3$ and set the default $r_t$ to 0.001.
Users can adjust $r_t$ from our graphical user interface to change the amount of zooming.
We compare a rendered frame with and without the zooming effect in~\figref{zoom}.

\begin{figure*}[t]
	\centering
	\begin{tabular}{cc}
		\includegraphics[width=0.48\textwidth]{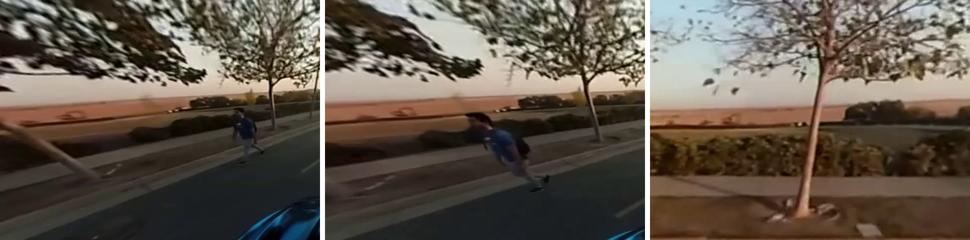} \hspace{-2mm} &
		\includegraphics[width=0.48\textwidth]{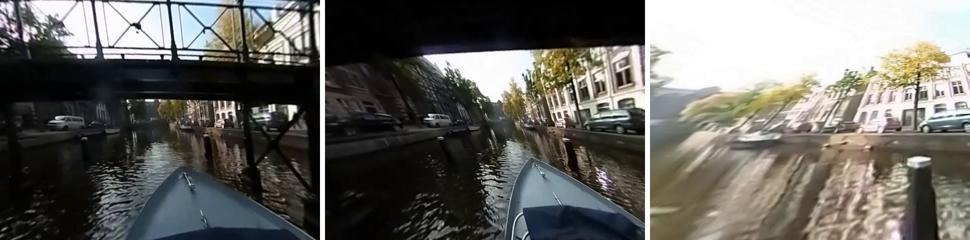} 
		\\
		\includegraphics[width=0.48\textwidth]{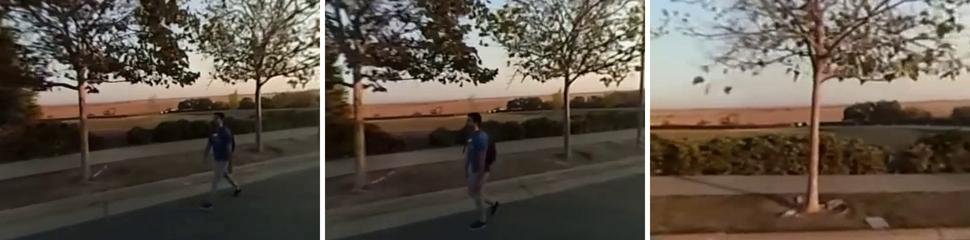} \hspace{-2mm} &
		\includegraphics[width=0.48\textwidth]{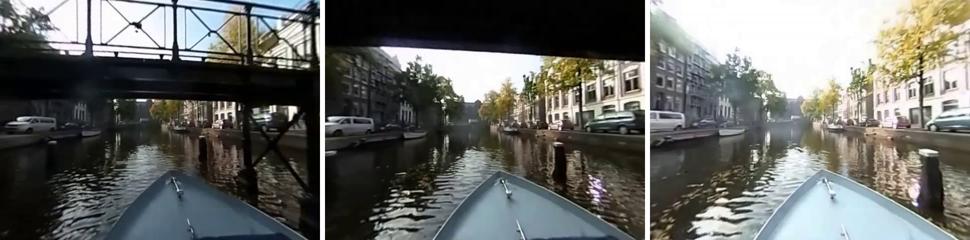} 
		\\
		\textsc{Campus} Video
		\hspace{-2mm} &
		\textsc{Boat} Video
	\end{tabular}
	\vspace{-2mm}
	\caption{Results after 2D stabilization. 
		\textbf{Top}: stabilized frames based on homographies. 
		\textbf{Bottom}: stabilized frames by the adaptive model selection method.
		The adaptive video stabilization approach selects the best motion models from translation, similarity transformation and homography according to the AIC~\protect\cite{AIC}. 
		Our method effectively reduces temporal artifacts (e.g., shearing or wobbling).
	}
	\vspace{-2mm}
	\label{fig:adaptive_stabilization}
\end{figure*}

{\flushleft \bf{Adaptive video stabilization.}}
%
Existing 2D video stabilization methods~\cite{Liu-TOG-2013,Joshi-TOG-2015} estimate the homography from matched features between consecutive frames.
However, we find that the frame-to-frame homographies may not be estimated well when there exist large camera panning, zooming, or rapid scene changes (e.g., from indoor to outdoor scenes).
Such scenarios commonly occur in our hyperlapses, and thus the stabilization approach based on the homographies does not perform well.
On the other hand, it is not effective to simply use translation or similarity transformation to reduce jitters in the videos.
In this work, we adopt a simple yet effective stabilization method to adaptively select the suitable motion model in each frame.

%
We first extract a set of feature trajectories using the Harris corners and Brief descriptors~\cite{Brief} on the NFOV video.
For each frame, we use the RANSAC method~\cite{RANSAC} to compute three motion models: homography, similarity transformation, and pure translation.
We then select the best motion models for each frame using the AIC~\cite{AIC}.
%
The AIC is derived from the information theory to measure the reliability of an estimated model based on the residual and number of parameters, which is defined as:
\begin{equation}
AIC = n \ln\left( \frac{\sum_{i=1}^n r_i^2}{n} \right) + 2k,
\end{equation}
where $r_i$ is the re-projection error of feature points as computed in~\eqnref{reprojection_cost}, $n$ is the number of feature pairs, and $k$ is the number of the parameters in the motion model.
We compute the AIC values for the three motion models and select the one with the minimum AIC value for each frame.
%
%

To smooth the camera motion, we adopt the single-path scheme by Liu~\etal~\cite{Liu-TOG-2013}.
In order to describe the motion in consecutive frames in a video, we convert the 2D translation and similarity transformation into $3 \times 3$ matrices. 
Let $\mathbf{P}_t$ be the camera pose at frame $t$, which can be written as a sequence of transformations:
\begin{equation}
\mathbf{P}_t = \mathbf{H}_t \mathbf{P}_{t-1} = \mathbf{H}_t \mathbf{H}_{t-1} \cdots \mathbf{P}_0,
\end{equation}
where $\mathbf{H}_t$ is the transformation matrix that warps features in frame $t - 1$ to frame $t$ and $\mathbf{P}_0$ is the identity matrix.
We find the optimal camera poses $\bar{\mathbf{P}}$ by minimizing the following function:
\begin{equation}
E(\mathbf{\bar{P}}) = \sum_t\left( 
\| \mathbf{\bar{P}}_t - \mathbf{P}_t \|^2 
+ \lambda \sum_{r\in\Omega_t} w_{t,r}  \| \mathbf{\bar{P}}_t - \mathbf{\bar{P}}_r \|^2
\right).
\label{eq:stabilization}
\end{equation}
Since the objective function in \eqnref{stabilization} is quadratic, we use the Jacobi-based iterative method to solve it:
\begin{equation}
\mathbf{\bar{P}}_t^{(k+1)} = \frac{1}{\gamma}\mathbf{P}_t + \sum_{r\in\Omega_t} \frac{2 \lambda w_{t,r}}{\gamma} \mathbf{\bar{P}}_r^{(k)},
\label{eq:iterative}
\end{equation}
where $\gamma = 1 + 2 \lambda \sum_{r\in\Omega_t}w_{t,r}$, $w_{t,r} = \exp\left( -(t - r)^2 / \sigma_t^2 \right)$, $\Omega_t$ are neighborhoods at frame $t$ and $k$ is an iteration index.
We set $\lambda = 2$ and $\sigma_t = 2$. 
For~\eqnref{iterative}, we iterate 5 times to optimize the camera poses.
After determining the smoothed camera motion, we compute the transformation matrix for each frame by $\mathbf{B}_t = \mathbf{\bar{P}}_t \mathbf{P}_t^{-1}$ and warp the corresponding frames to generate the final hyperlapse.

\figref{adaptive_stabilization} shows sample results after the 2D stabilization.
The stabilized results using pure homographies have significant shearing artifacts when the motion models are not accurately estimated.
In contrast, our adaptive approach selects reliable motion models based on the residuals of feature matching and renders more stable results.

\begin{figure*}[t]
	\centering
	\begin{tabular}{ccccc}
		\includegraphics[width=0.19\textwidth]{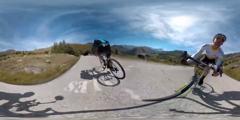} \hspace{-3.5mm} &
		\includegraphics[width=0.19\textwidth]{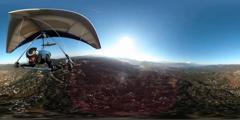} \hspace{-3.5mm} &
		\includegraphics[width=0.19\textwidth]{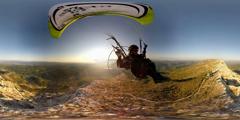} \hspace{-3.5mm} &
		\includegraphics[width=0.19\textwidth]{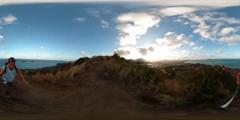} 	\hspace{-3.5mm} &
		\includegraphics[width=0.19\textwidth]{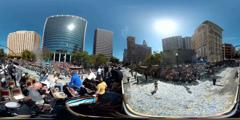}  \\
		\textsc{Biking}		\hspace{-3.5mm} &
		\textsc{Gliding 1}	\hspace{-3.5mm} &
		\textsc{Gliding 2}	\hspace{-3.5mm} &
		\textsc{Hiking}		\hspace{-3.5mm} &
		\textsc{Parade}		
		\\
		\includegraphics[width=0.19\textwidth]{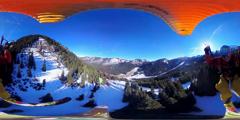} 	\hspace{-3.5mm} &
		\includegraphics[width=0.19\textwidth]{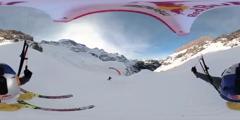} 	\hspace{-3.5mm} &
		\includegraphics[width=0.19\textwidth]{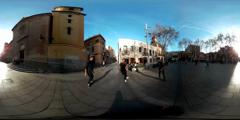} \hspace{-3.5mm} &
		\includegraphics[width=0.19\textwidth]{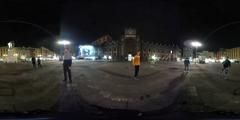} \hspace{-3.5mm} &
		\includegraphics[width=0.19\textwidth]{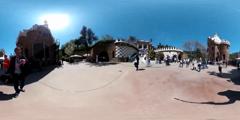} 
		\\
		\textsc{Ski 1}		\hspace{-3.5mm} &
		\textsc{Ski 2}		\hspace{-3.5mm} &
		\textsc{Soccer 1}	\hspace{-3.5mm} &
		\textsc{Soccer 2}	\hspace{-3.5mm} &
		\textsc{Walking}
		\\
		\includegraphics[width=0.19\textwidth]{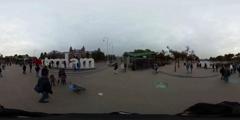} 	\hspace{-3.5mm} &
		\includegraphics[width=0.19\textwidth]{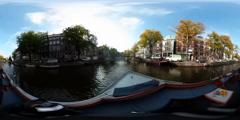} 			\hspace{-3.5mm} &
		\includegraphics[width=0.19\textwidth]{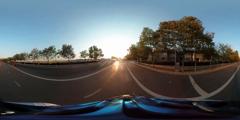} 		\hspace{-3.5mm} &
		\includegraphics[width=0.19\textwidth]{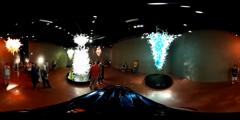} 		\hspace{-3.5mm} &
		\includegraphics[width=0.19\textwidth]{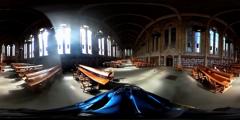} 	
		\\
		\textsc{Amsterdam}		\hspace{-3.5mm} &
		\textsc{Boat}			\hspace{-3.5mm} &
		\textsc{Campus}			\hspace{-3.5mm} &
		\textsc{Chihuly}		\hspace{-3.5mm} &
		\textsc{Library}	
		\\
		\includegraphics[width=0.19\textwidth]{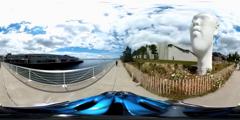} 		\hspace{-3.5mm} &
		\includegraphics[width=0.19\textwidth]{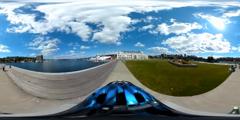} 		\hspace{-3.5mm} &
		\includegraphics[width=0.19\textwidth]{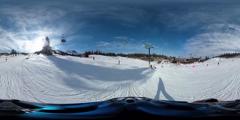}   	\hspace{-3.5mm} &
		\includegraphics[width=0.19\textwidth]{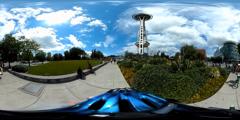} 	\hspace{-3.5mm} &
		\includegraphics[width=0.19\textwidth]{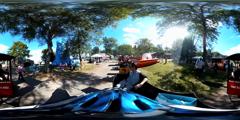} 
		\\
		\textsc{Park 1}			\hspace{-3.5mm} &
		\textsc{Park 2}			\hspace{-3.5mm} &
		\textsc{Snowboard}		\hspace{-3.5mm} &
		\textsc{Space Needle}	\hspace{-3.5mm} &
		\textsc{Train}
	\end{tabular}
	\vspace{-1mm}
	\caption{Thumbnails for videos in our datasets. The first two rows are videos from the Youtube dataset. The last two rows are videos from the Ricoh dataset.}
	\label{fig:datasets}
\end{figure*}

\section{Results and Discussions}
\label{sec:results}
We evaluate the proposed algorithm on a diverse set of $360^\circ$ videos from a range of activities and scenes.
The full input sequences and output hyperlapses are presented in the supplementary material.
Since our goal is to generate interesting and attractive hyperlapses, there is no objective metric for performance evaluation.
We thus conduct a large-scale user study to compare several variations of the proposed algorithm and existing methods.

\subsection{Performance}
%
We implement the proposed algorithm on a machine with an Intel Core-i7 3.4 GHz CPU and 32 GB RAM.
\tabref{time} shows the computational time of the proposed algorithm.
We break down the process into different stages and report the run-time per frame.
The main computational load lies in the optical flow estimation and the construction of TSPs. 
We note that the computed optical flow is also used in the focus of expansion estimation and spatial-temporal saliency detection stages.
The optical flow estimation can be further sped up by using CNN-based methods, e.g., FlowNet~\cite{flownet}.

\begin{table}
	\centering
	\caption{Computational time of each stage in our framework. We measure the per-frame run-time on a video with the spatial resolution of $1920 \times 960$.}
	\vspace{-1mm}
	\begin{tabular}{lc}
		\hline
		\textbf{Stage} & \textbf{Time / Frame} \\
		\hline\hline
		$360^\circ$ video stabilization 	& 0.480 s \\
		Optical flow estimation and TSP 	& 1.300 s \\
		Focus of expansion estimation 		& 0.053 s \\
		FCN semantic segmentation 			& 0.042 s \\
		Spatial-temporal saliency detection & 0.021 s \\
		Camera view planning 				& 0.002 s \\
		Saliency-aware frame selection		& 0.016 s \\
		Path refinement and rendering		& 0.010 s \\
		\hline
		Total 								& 1.924 s \\
		\hline
	\end{tabular}
	\vspace{-2mm}
	\label{tab:time}
\end{table}

\begin{table*}
	\centering
	\footnotesize
	\caption{
		Detailed video properties and semantic labels used in the ROI detection for the Youtube and Ricoh datasets.
		The fourth and ninth columns list the semantic labels used in \textit{Ours-saliency} method.
		The fifth and tenth columns list the semantic labels used in \textit{Ours-selected} method.
	}
	\vspace{-0.1cm}
	\begin{tabular}{ccccc|ccccc}
		\hline
		\multirow{2}{*}{\textcolor{blue}{Video}} & 
		\multirow{2}{*}{\textcolor{blue}{Length}} &
		\multirow{2}{*}{\textcolor{blue}{Speed-up}} &
		{\textcolor{blue}{Labels}} &
		{\textcolor{blue}{Labels}} &
		\multirow{2}{*}{
			\textcolor{blue}{Video}} & 
		\multirow{2}{*}{
			\textcolor{blue}{Length}} &
		\multirow{2}{*}{
			\textcolor{blue}{Speed-up}} &
		{\textcolor{blue}{Labels}} &
		\textcolor{blue}{Labels} 
		\\
		& & & \textcolor{blue}{(\textit{saliency})} & \textcolor{blue}{(\textit{selected})} &
		& & & \textcolor{blue}{(\textit{saliency})} & \textcolor{blue}{(\textit{selected})}
		\\
		\hline\hline
		\textcolor{Brown}{Biking}          & 3:20 & 4$\times$  & tree     & person                 & 
		\textcolor{Brown}{Boat}            & 6:40 & 12$\times$ & boat     & building               \\
		\textcolor{Brown}{Soccer 1}        & 0:30 & 4$\times$  & ground   & person                 & 
		\textcolor{Brown}{Amsterdam}       & 4:40 & 8$\times$  & tree     & building               \\
		\textcolor{Brown}{Soccer 2}        & 0:50 & 4$\times$  & ground   & person                 & 
		\textcolor{Brown}{Chihuly}         & 5:33 & 9$\times$  & tree     & light, pottedplant     \\
		\textcolor{Brown}{Gliding 1}       & 4:47 & 8$\times$  & car      & aeroplane, mountain    & 
		\textcolor{Brown}{Park 1}          & 2:46 & 8$\times$  & tree     & building, person       \\
		\textcolor{Brown}{Gliding 2}       & 3:08 & 4$\times$  & grass    & mountain               & 
		\textcolor{Brown}{Park 2}          & 2:06 & 8$\times$  & ground   & boat, building         \\
		\textcolor{Brown}{Parade}          & 5:00 & 8$\times$  & building & tree                   & 
		\textcolor{Brown}{Space Needle}    & 5:00 & 8$\times$  & tree     & aeroplane, bicycle     \\
		\textcolor{Brown}{Ski 1}           & 2:28 & 4$\times$  & sky      & person                 & 
		\textcolor{Brown}{Train}           & 5:00 & 8$\times$  & sky      & car                    \\
		\textcolor{Brown}{Ski 2}           & 0:42 & 4$\times$  & snow     & mountain, boat         & 
		\textcolor{Brown}{Campus}          & 4:00 & 6$\times$  & tree     & aeroplane, building    \\
		\textcolor{Brown}{Hiking}          & 1:56 & 8$\times$  & sky      & person, mountain       & 
		\textcolor{Brown}{Snowboard}       & 6:40 & 4$\times$  & snow     & person                 \\
		\textcolor{Brown}{Walking}         & 1:10 & 4$\times$  & tree     & person                 & 
		\textcolor{Brown}{Library}         & 1:56 & 8$\times$  & wall     & window                 \\
		\hline
	\end{tabular}
	\label{tab:dataset}
	\vspace{-1mm}
\end{table*}

\subsection{Datasets}
\label{sec:dataset}
%
We capture first-person $360^\circ$ videos using the Ricoh Theta S camera. 
The camera is mounted on a helmet or a tripod, and videos are recorded during walking, bicycling, boating or skiing.
We use 10 videos as our Ricoh dataset for experiments. 
In addition, we select 10 videos from the $360^\circ$ video dataset by Su~\etal~\cite{pano2vid}, which are collected from Youtube.
We choose the videos without shot boundaries since our algorithm tracks features across the entire videos without considering shot changes.
These videos are captured from different $360^\circ$ cameras, and the scenes are significantly different from the Ricoh dataset.
The image resolutions are $1920 \times 960$ and $1280 \times 640$ pixels in the Ricoh and Youtube datasets, respectively.
\figref{datasets} presents a sample frame from each sequence, and Table~\ref{tab:dataset} shows the properties of these two datasets.

\subsection{Graphical User Interface}
\label{sec:GUI}

\begin{figure*}[t]
	\centering
	\begin{tabular}{cc}
		\includegraphics[height=0.35\textwidth]{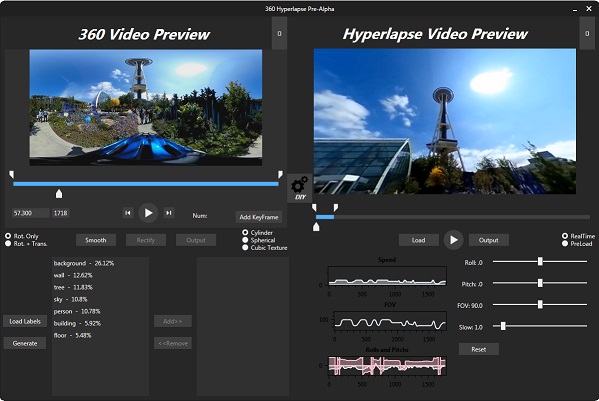} &
		\includegraphics[height=0.35\textwidth]{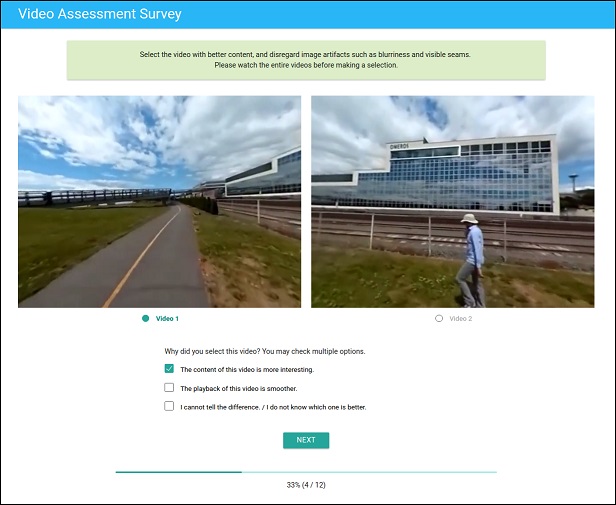} 
		\vspace{1mm}\\
		(a) Graphical user interface of our system & 
		(b) Interface of our user study
	\end{tabular}
	\vspace{-1mm}
	\caption{Screenshots of our GUI and user study.
		(a) We develop a GUI that integrates all the functions and allows users to choose the semantic labels for customizing hyperlapses.
		(b) We conduct the pairwise comparison that forces users to choose a preferred video. We also ask users to choose the reasons why they prefer the selected video.
	}
	\label{fig:GUI_userstudy}
	\vspace{-3mm}
\end{figure*}

We develop a graphical user interface (GUI) that integrates all the functions for rendering hyperlapses.
%
%
Once an input $360^\circ$ video is stabilized, our algorithm generates a list of semantic labels based on the video contents.
%
Each user can select multiple labels of interest and adjust the default FOV or speed-up rate to generate the hyperlapse of the individual preference. 
An optimal camera path is determined, and a hyperlapse is generated by the proposed algorithm. 
The rendered hyperlapse and video information (e.g., per-frame speed, FOV and viewing direction) can be viewed on the GUI (\figref{GUI_userstudy}(a)).

\subsection{Experimental Comparisons}
%
As generating hyperlapses from 360 videos is a new application, there are no direct competing methods.
We compare the proposed algorithm with the Pano2Vid method~\cite{pano2vid} which is developed for generating NFOV videos from $360^\circ$ videos.
The results of Pano2Vid are kindly provided by the authors.
%
%
Since the Pano2Vid method does not accelerate the videos, we use the hyperlapse algorithm by Joshi~\etal~\cite{Joshi-TOG-2015} to achieve the same speed-up rate as the proposed algorithm.

Our algorithm generates a camera path from the detected ROIs and semantic labels.
By default, the label with the maximum cumulative saliency score is selected.
We denote the proposed algorithm with the default settings as {\em Ours-saliency}.
However, the semantic labels with higher saliency scores may come from background-like objects (e.g., ground, sky or tree).
The hyperlapses rendered from such labels may not be interesting to people.
Therefore, we manually choose some interesting labels (e.g., building, aeroplane, and person) for detecting ROIs, and we denote the proposed method with this setting as {\em Ours-selected}.
Our algorithm can also generate results without using any ROIs or semantic labels.
In such cases, the path is mainly determined by the focus of expansion, and the viewing direction in the hyperlapses remains forward-looking without any panning.
We denote the proposed algorithm with this baseline setting as {\em Ours-forward}.

We summarize the video length, target speed-up rate, and the labels used in our method for each sequence in~\tabref{dataset}.
We compare with Pano2Vid on the Youtube dataset and compare the three variations of our method on all 20 videos.
We note that the results of Pano2Vid have a fixed FOV of $65.5^\circ$.
For fair comparisons, we set the output FOV of our results on the Youtube dataset to $65.5^\circ$ as well.
For videos on the Ricoh dataset, we set the default FOV to $100^\circ$.
All the videos are available on our project website at~\url{http://vllab.ucmerced.edu/wlai24/360hyperlapse}.

\subsection{User study}
\label{sec:userstudy}
We conduct a large-scale user study based on pairwise comparisons that require each participant to choose a preferred sequence from a pair of rendered videos.
We design a web interface that allows users to watch two videos at the same time, as shown in~\figref{GUI_userstudy}(b).
The users can also replay these videos individually to help them make decisions.
Every participant is required to evaluate results on 20 different input videos.
We randomly choose the results from the compared methods for each sequence, and the videos are displayed in a random order.
Each subject is required to watch the entire videos before making a decision.
%
We note that the $360^\circ$ videos from the Youtube dataset are captured from different cameras, and some videos contain visible stitching seams or have low resolution.
Therefore, we ask the subjects to disregard artifacts such as blurriness and visible seams when evaluating the rendered hyperlapses.
%

To remove careless evaluations from casual users, we include two video pairs for the sanity check.
We synthesize paths where the cameras keep pointing to the sky or ground.
These videos are not interesting at all when compared to those generated by the evaluated methods.
We discard the results if a participating subject fails the sanity check more than once or finishes the study in haste (e.g., less than 5 minutes).

\begin{figure*}
	\centering
	\includegraphics[width=0.5\textwidth]{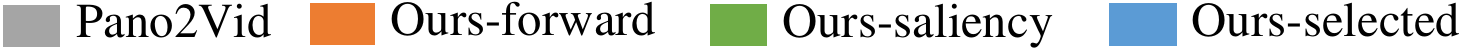} \vspace{2mm}\\
	\begin{tabular}{ccc}
		\includegraphics[width=0.31\textwidth]{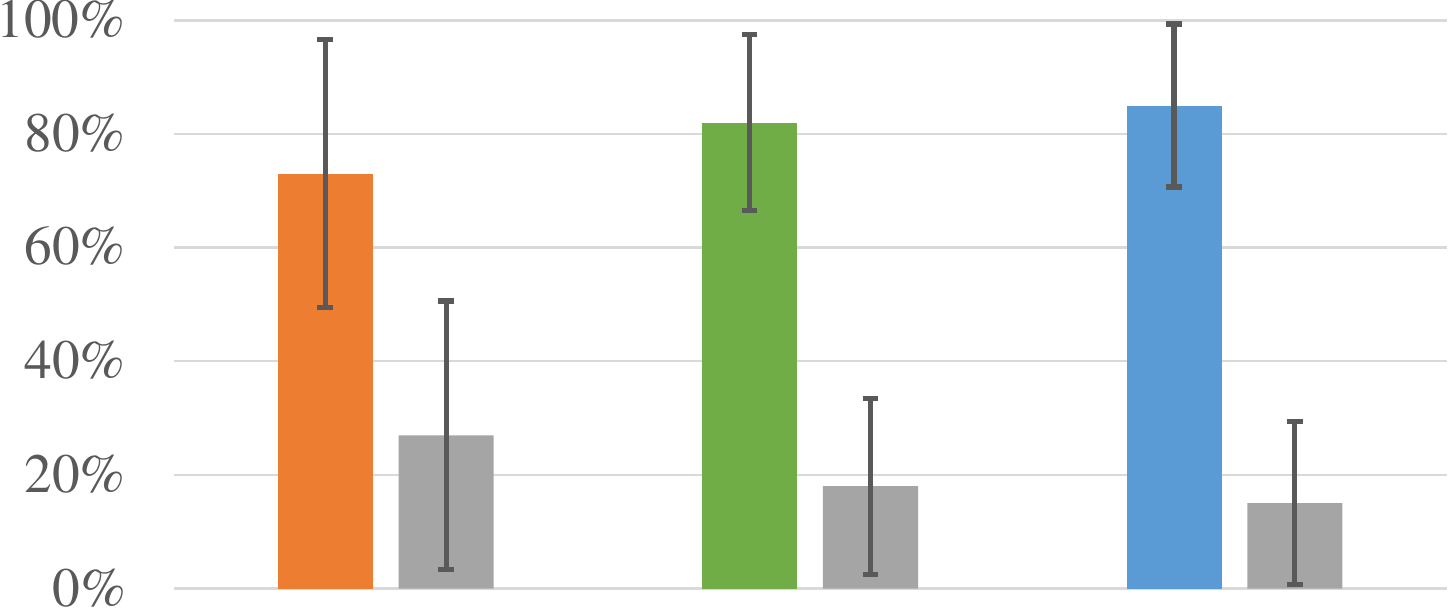} &
		\includegraphics[width=0.31\textwidth]{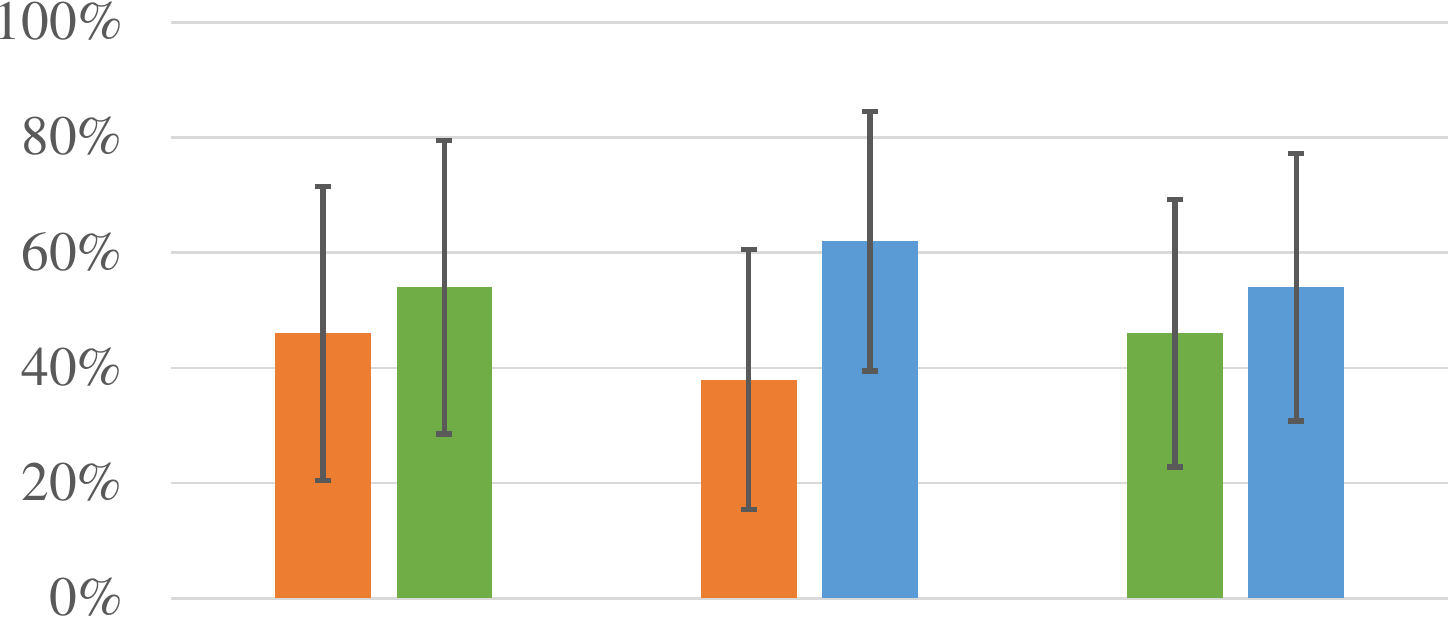} &
		\includegraphics[width=0.31\textwidth]{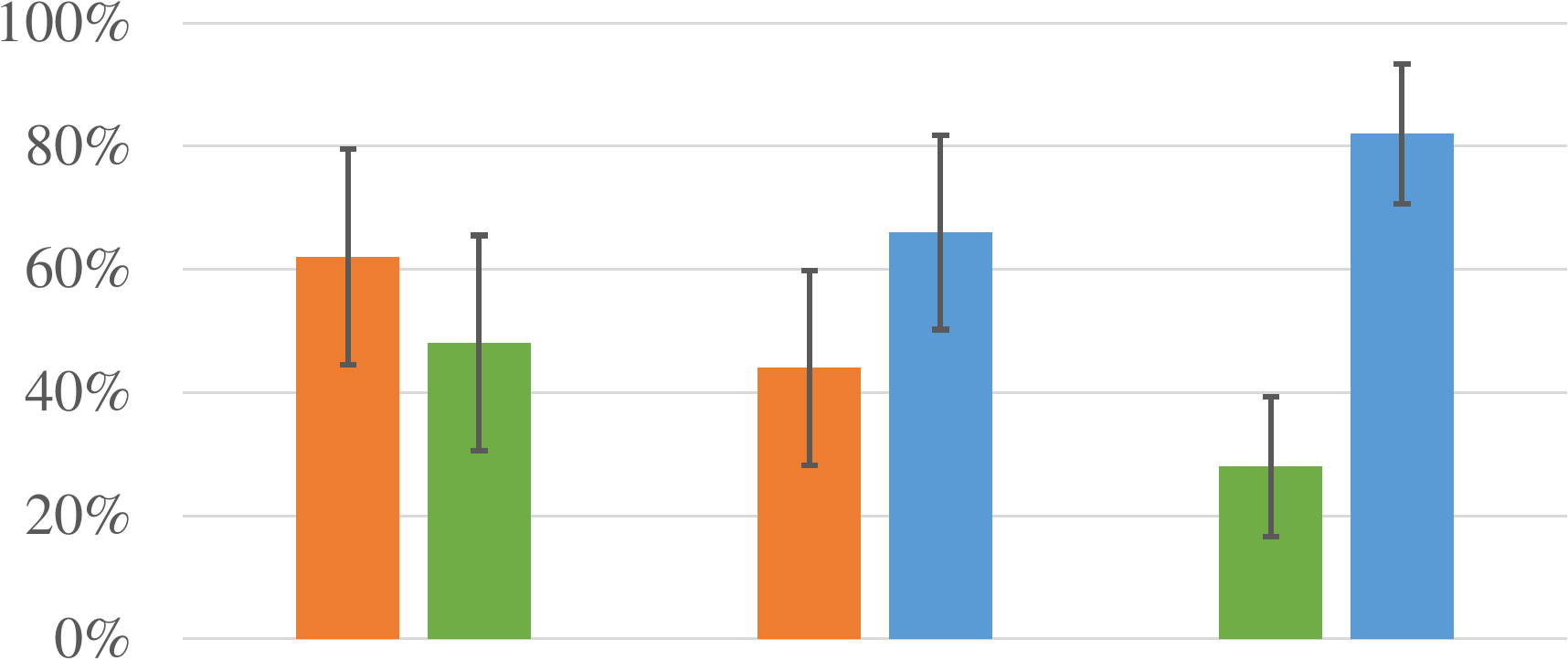} \\
		(a) & (b) & (c)
	\end{tabular}
	\vspace{-2mm}
	\caption{Percentage of obtained votes from the user study.
		(a) Pano2Vid v.s. our methods on the Youtube dataset.
		(b) Comparisons of our variations on the Youtube dataset.
		(c) Comparisons of our variations on the Ricoh dataset.}
	\label{fig:votes}
	\vspace{2mm}
\end{figure*}

We obtain the results from 200 participants using Amazon Mechanical Turk.
%
Since we randomly choose pairs of videos for each evaluation, some videos may not be selected for the same numbers of time.
We thus balance the subject votes by uniformly sampling the results such that each video pair is compared at the same frequency.
%
In~\figref{votes}(a), we show the percentage of obtained votes for our methods and Pano2Vid on the Youtube dataset.
We note that the virtual cameras by the Pano2Vid method often fixate at the bottom or sides of the scenes (as shown in the supplementary videos).
The results from our algorithm are preferred over those by the Pano2Vid method 
as we consider the direction of camera forward motion (i.e., FOE), which is crucial for providing a better understanding of scenes and activities in videos.


\begin{figure*}
	\centering
	\includegraphics[width=0.4\textwidth]{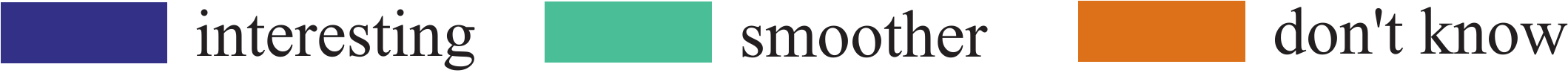} \vspace{2mm}\\
	\begin{tabular}{ccc}
		\includegraphics[width=0.32\textwidth]{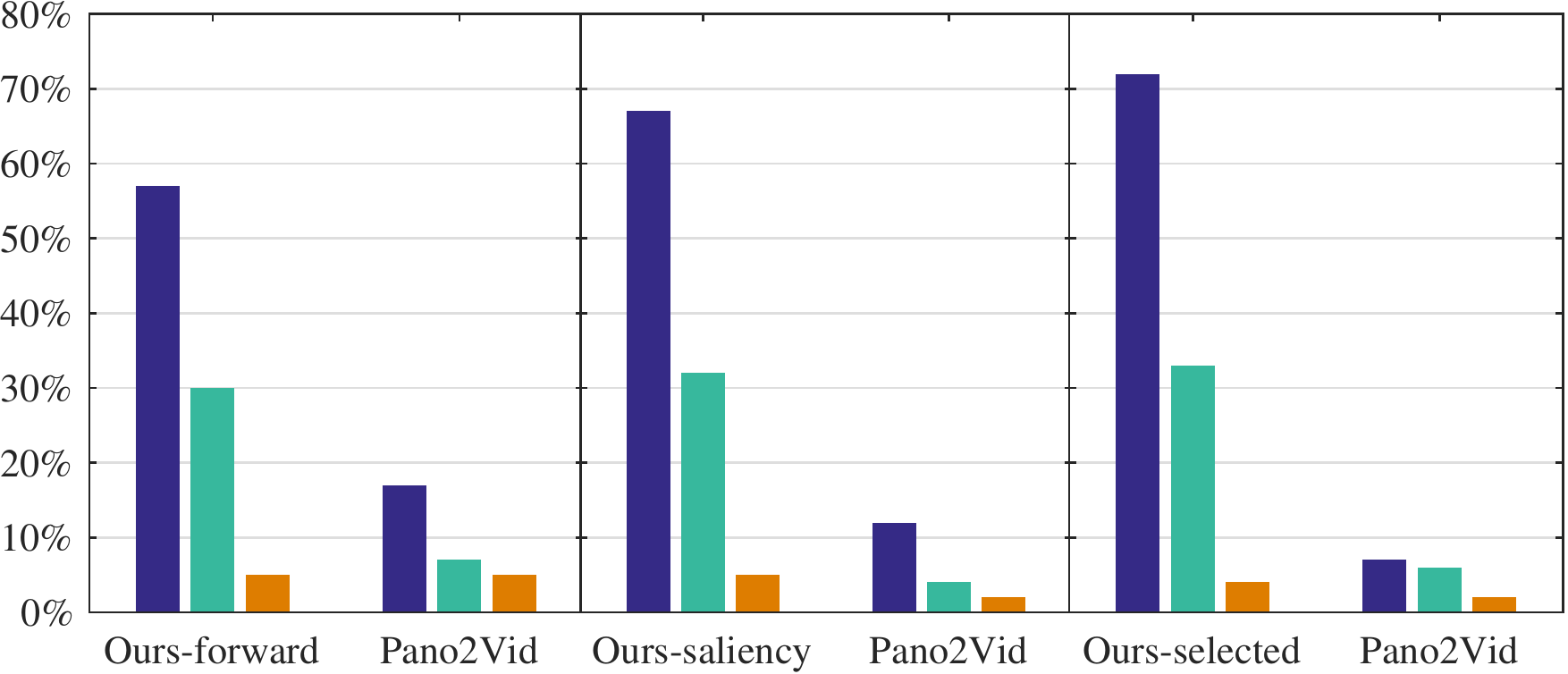} 
		\hspace{-3mm} &
		\includegraphics[width=0.32\textwidth]{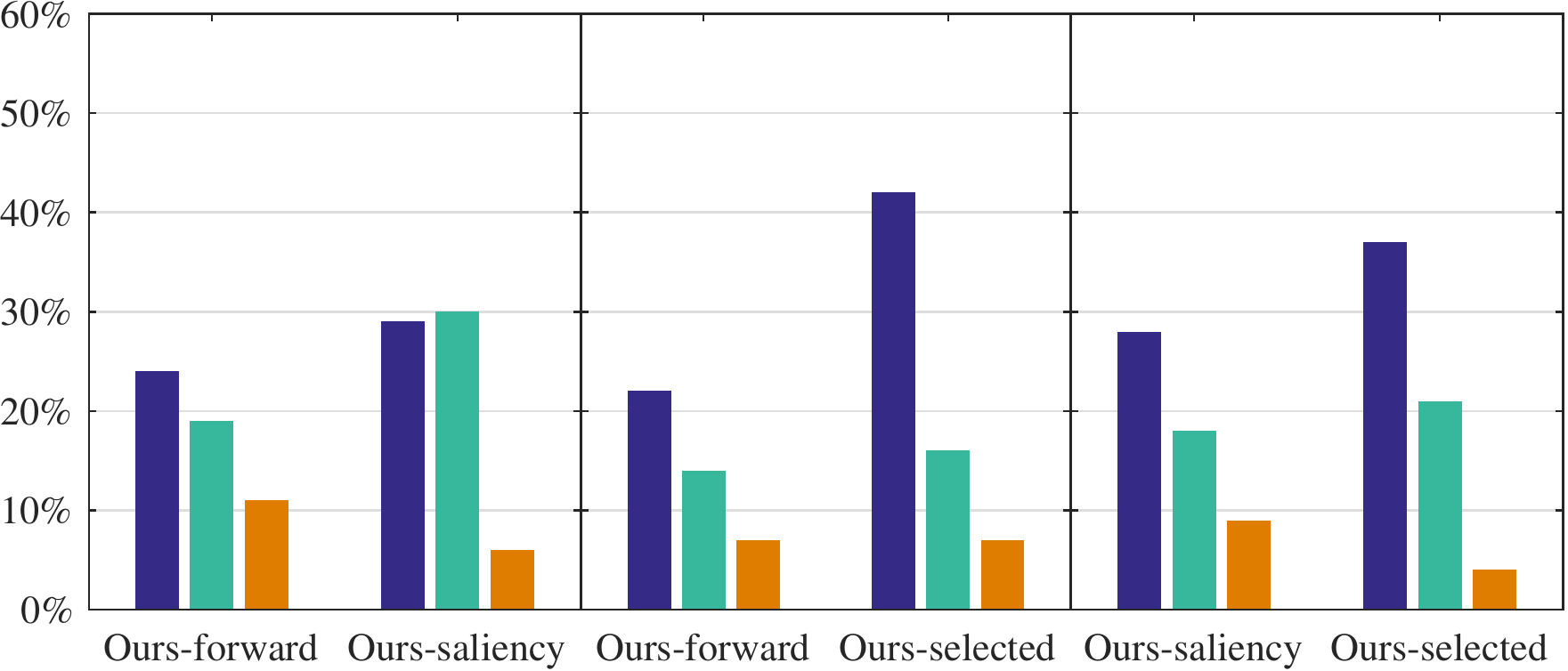} 
		\hspace{-3mm} &
		\includegraphics[width=0.32\textwidth]{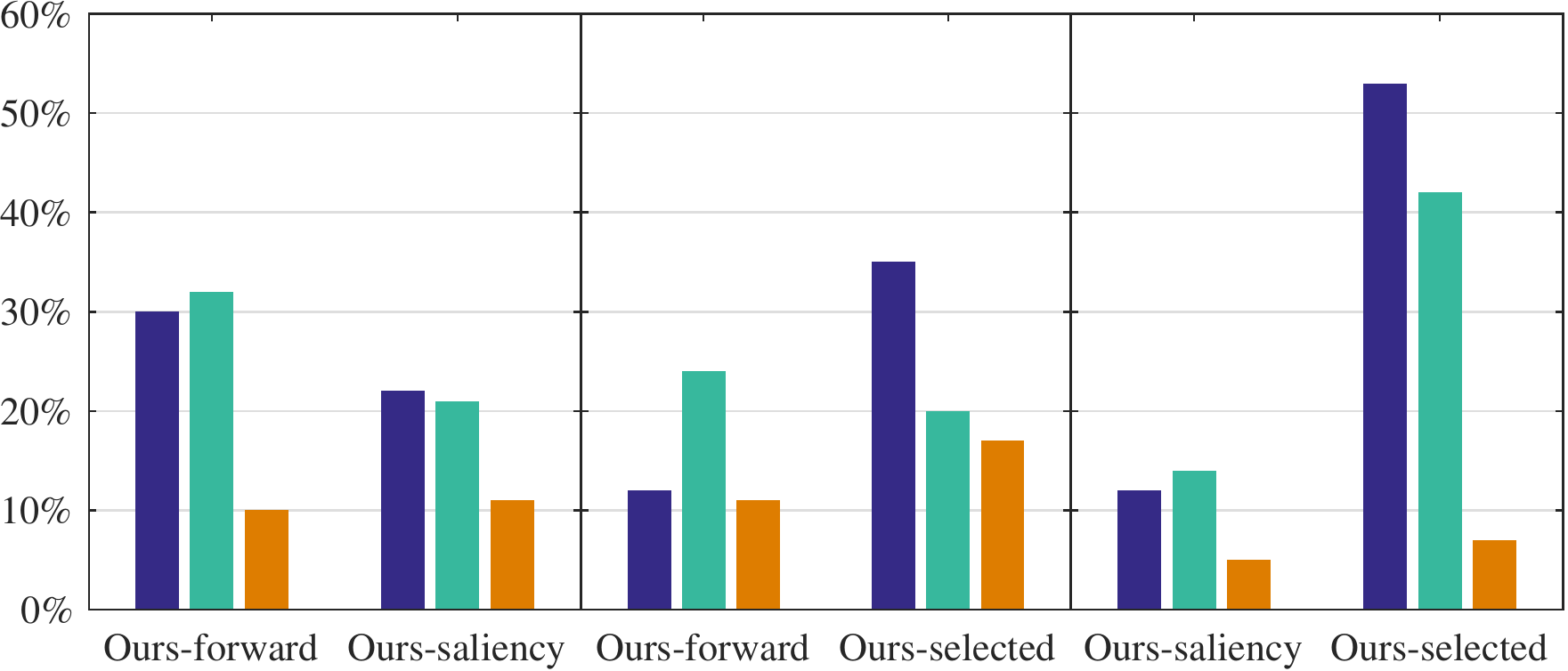} \\
		(a) 
		\hspace{-3mm} & 
		(b) 
		\hspace{-3mm} & 
		(c)
	\end{tabular}
	\vspace{-2mm}
	\caption{Percentage of reasons to favor one method over the other 
		from the human subject study.
		(a) Pano2Vid v.s. our methods on the Youtube dataset.
		(b) Comparisons of our variations on the Youtube dataset.
		(c) Comparisons of our variations on the Ricoh dataset.}
	\label{fig:reason}
	\vspace{2mm}
\end{figure*}

We further show the evaluations among the three variations of our algorithm in~\figref{votes}(b) and (c).
The results from \textit{Ours-selected} are preferred by participating subjects than those from \textit{Ours-saliency} since the semantic labels with higher saliency scores may not generate interesting hyperlapses if the detected ROIs are background-like objects.
Compared with the baseline method (\textit{Ours-forward}), more subjects prefer the results with camera panning to look at interesting objects.
However, we find that more users vote for \textit{Ours-forward} than \textit{Ours-saliency} on the Ricoh dataset and the numbers of votes are very close on the Youtube dataset.
This can be explained by the fact that the camera panning and speed change affect the smoothness of the generated videos.
If the cameras do not look at real interesting objects, users tend to prefer the video with smoother playback.

For each evaluation, we also ask participants to choose one of the reasons why they prefer the selected video:
%
\begin{itemize}
	\item ``The content of this video is more interesting."
	\item ``The playback of this video is smoother."
	\item ``I cannot tell the difference. / I do not know which one is better."
\end{itemize}
We note that even if users cannot tell the difference between two videos, they are required to choose one.
We present the percentage of chosen reasons in~\figref{reason}.
It is clear that not only the interestingness of the content but also the smoothness of playback are crucial to the perceived quality of videos.
The results by our approach are consistently preferred by users over those by the Pano2Vid method.

\subsection{Limitations}
%
Due to computational efficiency, we use image-based semantic segmentation with the spatial-temporal over-segmentation algorithm to segment each video.
As large objects are split into multiple small TSPs, the top few TSPs may come from the same object.
This, in turn, may affect the amount of zooming and reduce the chances to detect other interesting objects in the video.

Currently, the most preferred hyperlapses generated by our system are obtained through human-selected semantic labels.
We are interested in developing a fully automatic method that incorporates more high-level semantic and saliency detection (e.g., landmark detection).

Our algorithm does not handle generic wide-angle videos (i.e., less than $360^\circ$) as we project the video frames to a 3D sphere for stabilization.
For a generic wide-angle video, a camera calibration is required to estimate the projection parameters. 
However, we note that our view planning algorithm can be easily adapted to handle general videos by constraining the spatial range of the camera path.

%
%

\section{Conclusions}
In this work, we present a semantic-driven approach for generating hyperlapses from $360^\circ$ videos.
We analyze video contents by semantic segmentation as well as spatial-temporal saliency and identify interesting regions or objects to guide the camera path planning.
The proposed path planning algorithm can automatically determine a smooth camera path such that interesting objects can be viewed with variable speed.
We use the saliency-aware frame selection and adaptive 2D video stabilization to efficiently generate smooth hyperlapses.
%
Furthermore, our system also facilitates users to customize the hyperlapses by selecting individual preferences.
A large-scale user study shows that the proposed system performs favorably against the state-of-the-art method for generating hyperlapses from $360^\circ$ videos. 
%
%

\section*{Appendix A: $\mathbf{360^\circ}$ Video Stabilization}

\label{sec:360stabilization}
%
Most first-person videos are casually captured during walking, running or bicycling.
The raw input videos often contain significant camera shakes, twists and turns.
As a result, we stabilize the entire $360^\circ$ video before analyzing contents and planning camera paths.

We adopt a similar approach to~\cite{Kasahara-2015} for estimating and smoothing the relative rotation between consecutive video frames.
We assume that the relative translation between frames is negligible, and the frame-to-frame transformation between $360^\circ$ views can be described by 3D rotations (yaw, pitch, and roll).
First, we track a set of sparse 2D feature trajectories across multiple frames using the KLT tracker~\cite{KLT}.
Next, we convert the 2D feature points into 3D vectors on a unit sphere.
We use the method of~\cite{Horn-1987} to represent rotations with 4D unit quaternions and solve for the quaternions with a closed form solution.
We denote the quaternions from frame $t$ to $t + 1$ as $\mathbf{q}_t$.
To relate them to the first frame, we connect the quaternions with $\mathbf{Q}_t = \prod_{i=1}^t \mathbf{q}_i$.
We then use the quaternion interpolation (i.e., the \textit{Slerp} operation) with a Gaussian weight function to obtain the smoothed quaternions $\bar{\mathbf{Q}}_t$.
The width of the Gaussian kernel can be determined from our graphical user interface. 
After smoothing, we compute the warping transformation $\mathbf{R}_t = \mathbf{Q}_t^{-1} \bar{\mathbf{Q}}_t$ to warp the $i$-th frame to generate a stabilized $360^\circ$ video.

\begin{figure}
	\centering
	\hspace{-5mm}
	\includegraphics[width=1.0\columnwidth]{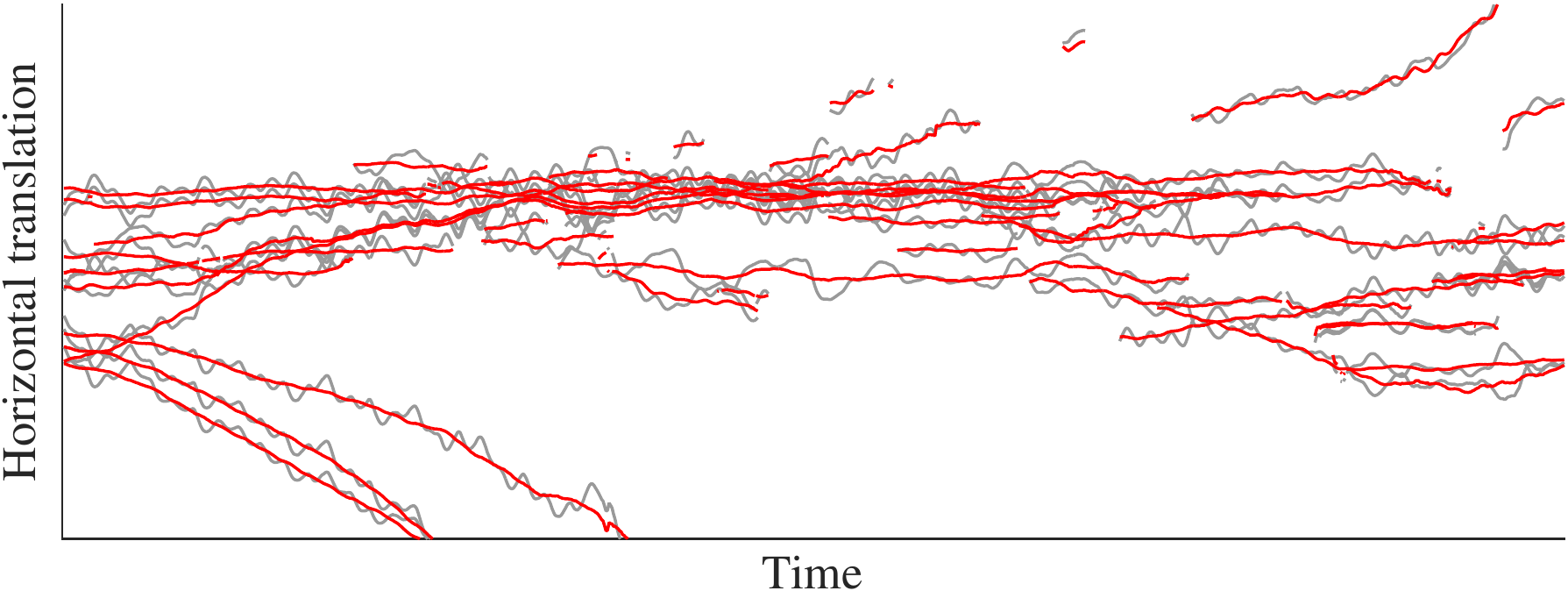}
	\caption{Feature trajectories before (grey curves) and after (red curves) stabilizing a $360^\circ$ video.}
	\label{fig:360_stabilization_trajectory}
\end{figure}

We illustrate the feature trajectories before and after the $360^\circ$ stabilization from the \textsc{Amsterdam} video in~\figref{360_stabilization_trajectory}.
The $360^\circ$ video stabilization effectively reduce the jittering in the input video.

\section*{Appendix B: Focus of Expansion Estimation}

\label{sec:FOE_estimation}
%
We detect several ROIs to control the panning of the virtual camera.
If no ROIs in a frame are detected, we need to use some motion prior 
to guide the camera direction.
Therefore, we estimate the focus of expansion as the prior for the camera path.

\begin{figure}
	\centering
	\begin{tabular}{cc}
		\multicolumn{2}{c}{\includegraphics[width=0.45\columnwidth]{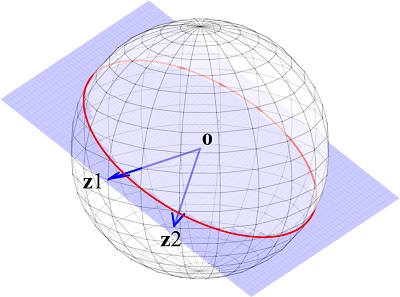}}
		\\
		\multicolumn{2}{c}{(a) Sphere and plane intersection}
		\vspace{1mm}\\
		\includegraphics[height=0.2\columnwidth]{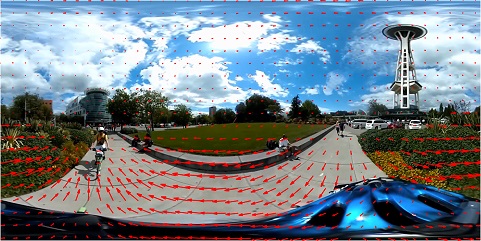} &
		\includegraphics[height=0.2\columnwidth]{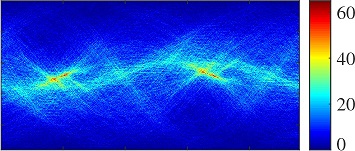} 
		\vspace{1mm}\\
		(b) Optical flow &
		(c) Votes from Hough transform
	\end{tabular}
	\caption{
		Intersection of a sphere and a plane passing through the sphere center defines a locus of points that are candidates of FOE and FOC.
		We use the Hough transform to estimate the FOE via optical flows.
	}
	\label{fig:FOE}
\end{figure}

The FOE is a single point from which all the optical flow vectors diverge.
The FOE indicates the direction of forward motion in a video.
On the other hand, the focus of contraction (FOC) is the antipodal point of FOE to which optical flow vectors converge.
The FOE and FOC can be parameterized as 2D points $(x, y)$ in the image coordinate.
We thus use the Hough transform with optical flows to estimate the locations of FOE and FOC.
Let $p_1$ and $v_1$ be an image point and its optical flow vector, respectively.
We find $p_2 = p_1 + v_1$ and project $p_1$ and $p_2$ to the 3D spherical coordinate.
Let $\mathbf{z}_1$ and $\mathbf{z}_2$ be the corresponding 3D vectors of $p_1$ and $p_2$, respectively.
%
Then, $\mathbf{z}_1$, $\mathbf{z}_2$ and the center of the unit sphere $\mathbf{o}$ form a plane and intersect with the unit sphere on a great circle, as illustrated in \figref{FOE}(a).
All the points on this great circle are candidates of FOE and FOC.
We note that the points on the great circle can be computed from rotating the vector $\mathbf{z}_1$ on the plane $\mathbf{o}\mathbf{z}_1\mathbf{z}_2$.
We first construct an orthogonal basis $\mathbf{B} = [\mathbf{b}_1, \mathbf{b}_2, \mathbf{b}_3 ]$ where:
\begin{align}
\mathbf{b}_1 &= \mathbf{z}_1, \nonumber\\
\mathbf{b}_2 &= \frac{\mathbf{z}_2 - (\mathbf{z}_1 \cdot \mathbf{z}_2)\mathbf{z}_1}{ \| \mathbf{z}_2 - (\mathbf{z}_1 \cdot \mathbf{z}_2)\mathbf{z}_1 \| }, \nonumber\\
\mathbf{b}_3 &= \mathbf{z}_1 \times \mathbf{z}_2. \nonumber
\end{align}
We then define a rotation matrix:
\begin{equation}
\mathbf{R}_{\theta} = 
\begin{pmatrix}
\cos\theta & -\sin\theta & 0 \\
\sin\theta & \cos\theta & 0 \\
0 & 0 & 1
\end{pmatrix}.
\end{equation}
The points on the great circle can be computed by:
\begin{equation}
\mathbf{y}_{\theta} = \mathbf{B} \mathbf{R}_{\theta} \mathbf{B}^{-1} \mathbf{z}_1,
\label{eq:foe_rotate}
\end{equation}
where $\theta \in [0^\circ, 360^\circ]$.
The physical meaning of~\eqnref{foe_rotate} is performing the basis change and then rotating $\mathbf{z}_1$ about the plane normal $\mathbf{b}_3$.

To determine the FOE and FOC in a frame, we construct a 2D voting matrix with the same size as the input image and re-project the 3D points $\mathbf{y}_{\theta}$ on the great circle back to the 2D image plane.
We aggregate the votes from all the optical flow vectors and find a pair of antipodal points that have the highest votes.
\figref{FOE}(c) shows an example of the voting matrix.
It is clear that two local maximums, which correspond to the FOE and the FOC, can be identified. 
The FOE and FOC can be distinguished by checking the direction of nearby flow vectors.

The above procedure can locate the FOEs in an input video.
However, the results are usually noisy since we process each frame independently.
We then apply a Gaussian filter to smooth the FOEs and obtain a stable curve, which is used as motion prior in camera path planning.

%



\bibliographystyle{IEEEtran}
\bibliography{hyperlapse}
%

%

\begin{IEEEbiography}[{\includegraphics[width=1in,height=1.25in,clip,keepaspectratio]{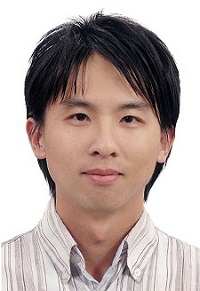}}]{Wei-Sheng Lai}
	is a Ph.D. candidate of Electrical Engineering and Computer Science at the University of California, Merced, CA, USA. He received the B.S. and M.S. degree in Electrical Engineering from the National Taiwan University, Taipei, Taiwan, in 2012 and 2014 respectively. His research interest includes computer vision, computational photography and deep learning.
\end{IEEEbiography}

\vspace{-0cm}

\begin{IEEEbiography}[{\includegraphics[width=1in,height=1.20in,clip,keepaspectratio]{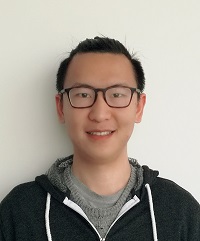}}]{Yujia Huang}
	is a Master student in Computer Vision at the Robotics Institute, School of Computer Science, Carnegie Mellon University, PA, USA. He received the B.S. degree in Computer Engineering from the University of Minnesota - Twin Cities, MN, USA in 2015. His research interest includes 3D computer vision, deep learning and robotics.
\end{IEEEbiography}

\vspace{-0cm}

\begin{IEEEbiography}[{\includegraphics[width=1in,height=1.25in,clip,keepaspectratio]{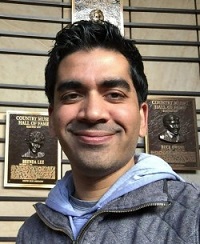}}]{Neel Joshi}
	is a Senior Researcher at Microsoft Research.  His work is in computer vision, computer graphics, and human-computer interaction, focusing particularly on imaging, computational photography, and creative tools for visual arts.  Neel holds an Sc.B. from Brown University, an M.S. from Stanford University, and a Ph.D. from U.C. San Diego all in Computer Science. He has held internships at Mitsubishi Electric Research Labs, Adobe Systems, and Microsoft Research, was a visiting professor at the University of Washington, and has been at Microsoft Research since 2008. 
\end{IEEEbiography}

\vspace{-0cm}

\begin{IEEEbiography}[{\includegraphics[width=1in,clip,keepaspectratio]{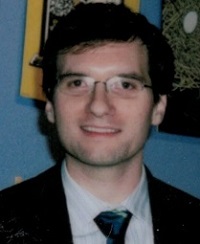}}]{Chris Buehler}
	is a Principal Engineering Manager in the Artificial Intelligence and Research (AI$\&$R) division at Microsoft. He received a Ph.D. in computer science from the Massachusetts Institute of Technology.
\end{IEEEbiography}

\vspace{-0cm}

\begin{IEEEbiography}[{\includegraphics[width=1in,height=1.25in,clip,keepaspectratio]{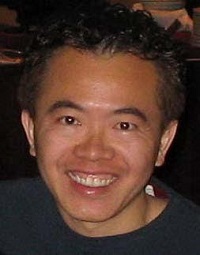}}]{Ming-Hsuan Yang}
	is a Professor of Electrical Engineering and Computer Science at the University of California, Merced, CA, USA. He received the Ph.D. degree in computer science from the University of Illinois at Urbana-Champaign, USA, in 2000. 	Yang received the NSF CAREER Award in 2012, and the Google Faculty Award in 2009. He is a senior member of the IEEE and ACM.
\end{IEEEbiography}

\vspace{-0cm}

\begin{IEEEbiography}[{\includegraphics[width=1in,height=1.25in,clip,keepaspectratio]{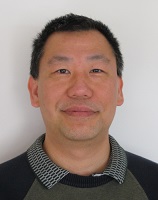}}]{Sing Bing Kang}
	is a Principal Researcher at Microsoft Research. His main area of interest is computational photography. Sing Bing has co-authored two books, Image-Based Rendering and Image-Based Modeling of Plants and Trees, and co-edited two others, Panoramic Vision and Emerging Topics in Computer Vision. He has served as area chair for the major computer vision conferences (ICCV, CVPR, ECCV) as well as papers committee member for SIGGRAPH and SIGGRAPH Asia. Sing Bing was program chair for ACCV 2007 and CVPR 2009. In addition, he was Associate Editor-In-Chief for IEEE TPAMI from 2010-2014. He is currently workshop co-chair for ICCV 2017 and ACVPR book series editor for Springer-Verlag. Sing Bing received his PhD in robotics from Carnegie Mellon University, and is an IEEE Fellow.
\end{IEEEbiography}




\end{document}